\documentclass[12pt]{article}
\usepackage{times}
\newenvironment{summarypara}%
  {\list{}{\leftmargin0.2cm\rightmargin=0.2cm}\bf \item[]}%
  {\endlist}

\usepackage{chngcntr}
\usepackage{graphicx,amsmath,amsfonts,amssymb}
\usepackage{booktabs}
\usepackage[utf8]{inputenc} % allow utf-8 input
\usepackage[T1]{fontenc}    % use 8-bit T1 fonts
\usepackage{hyperref}       % hyperlinks
\usepackage{microtype}      % microtypography
\usepackage[footnotesize]{caption}
\usepackage{lineno}
\usepackage{bbm}
\usepackage{xcolor}

\usepackage{latexsym}
\usepackage{paralist}
\usepackage{stmaryrd}
\usepackage{tikz}
\usepackage{titling}

\def\xprime{x_{m+1}}
\def\yprime{y_{m+1}}
\def \ll {$\llbracket$}
\def \rr {$\rrbracket$}
\def \red {\tikz\draw[red,fill=red] (0,0) circle (.8ex); }
\def \green {\tikz\draw[green,fill=green] (0,0) circle (.8ex); }
\def \yellow {\tikz\draw[yellow,fill=yellow] (0,0) circle (.8ex); }
\def \blue {\tikz\draw[blue,fill=blue] (0,0) circle (.8ex); }

\title{{\LARGE \textbf{More accurate behavioral predictions with hybrid Bayesian-connectionist models}}}

\date{}
\author{ \large \textbf{Brenden M. Lake}$^{1\ast}$, \textbf{Akshay K. Jagadish}$^{2}$, and \textbf{Guangyuan Jiang}$^{3}$ \\ 
\normalsize{$^1$Departments of Computer Science and Psychology, Princeton University} \\
\normalsize{$^2$Princeton AI Lab, Princeton University} \\
\normalsize{$^3$Department of Brain and Cognitive Sciences, Massachusetts Institute of Technology} \\
\normalsize{$^\ast$To whom correspondence should be addressed; E-mail:  brenden@princeton.edu}
}

\begin{document}
\baselineskip24pt
\maketitle

% \linenumbers
\begin{summarypara}
Researchers must often choose between Bayesian or neural network models of behavior, two paradigms with complementary strengths and weaknesses. An ideal paradigm would facilitate testing many kinds of representations and inductive biases; Bayesian models make this easy, while neural networks do not. Similarly, an ideal paradigm would avoid over-simplifications; neural networks make this easy, while Bayesian models do not. Here, we introduce Bayesian distillation with Behavioral Tuning (BBT) as an approach to getting the best of both traditions. BBT offers a simple recipe for model building: first, a neural network is trained to mimic a Bayesian model through synthetic data, and second, the network is fine-tuned on human behavior to capture additional structure and nuance. Across four case studies in human concept learning, we find that BBT outperforms traditional approaches at predicting human behavior while also revealing psychological insights, resulting in models that can both mimic Bayesian priors and capture heuristics and biases that violate simple modeling assumptions.
\end{summarypara}

\begin{figure}
\centering
\includegraphics[width=5in]{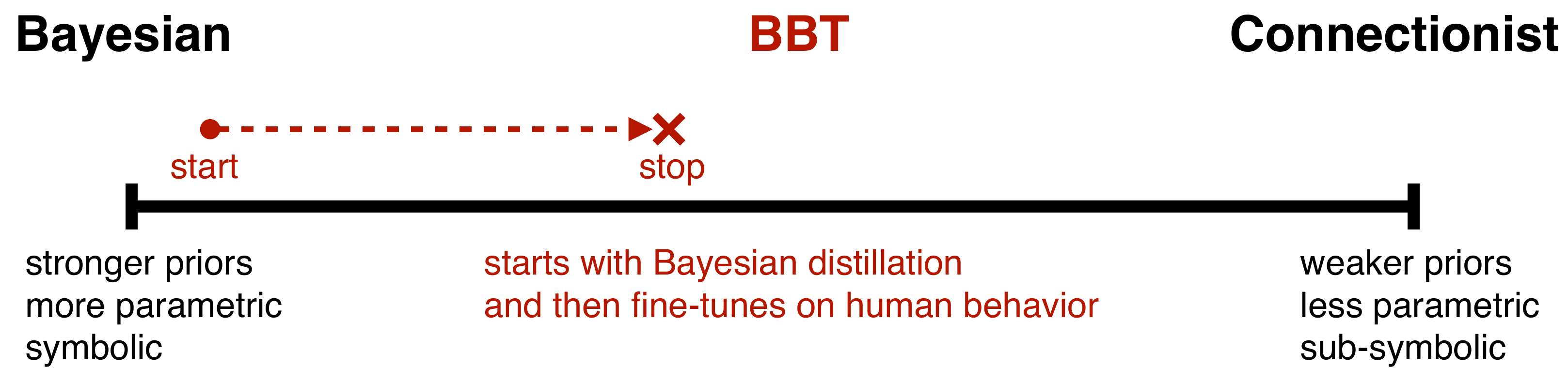}
\caption{The Bayesian and connectionist paradigms occupy opposite ends of a spectrum. Our Bayesian distillation with Behavioral Tuning (BBT) approach moves along this spectrum in two stages. First, a neural network is pretrained to mimic a Bayesian model through Bayesian distillation (marked as ``start'' in the figure). The network is then fine-tuned on human behavioral data, shifting it towards the connectionist end of the spectrum until performance on held-out data plateaus (marked as ``stop''). Depending on the nature of the behavioral data, the model may stop anywhere along the spectrum. Note that BBT is a model building method and its algorithmic steps are not a hypothesis about cognitive development.}
\label{fig_bbt_intro}
\end{figure}

\clearpage
% \linenumbers
Two major paradigms for cognitive modeling, Bayesian models and neural networks, have historically been in tension (see \cite{Griffiths2010a} and \cite{McClelland2010}, respectively, for contrasting reviews). The Bayesian tradition typically follows a top-down approach in the spirit of Marr's levels of analysis \cite{Marr1982}: researchers characterize a computational problem and its ideal solution, which is then compared to human behavior. Often, this involves formulating hypotheses as structured representations (rules, grammars, programs, etc.), priors as probabilistic beliefs over hypotheses, and learning as Bayesian updating of these beliefs \cite{Griffiths2010a,Tenenbaum2011}. The neural network (also known as connectionist) tradition follows a more emergentist approach \cite{McClelland2010a}: researchers construct systems of simple, interacting processing units and study the representations and behaviors that emerge through learning their connection weights, comparing these emergent behaviors to human behaviors \cite{PDPVol1,PDPVol2,Lecun2015}. Often, connectionists consider the kinds of symbolic representations commonplace in Bayesian models as approximate characterizations of more emergent, sub-symbolic processes \cite{McClelland2010}. Historically, this has led to two different views of the mind (Fig. \ref{fig_bbt_intro}): one emphasizing symbolic representations and strong inductive biases, and one emphasizing sub-symbolic representations and more generic learning mechanisms \cite{Griffiths2010a,McClelland2010}. 

These two views have also led to distinct modeling paradigms and toolkits. It is challenging to build models that are not strictly one kind or the other, or that achieve a combination of both paradigms' strengths, leading to tradeoffs. A relative advantage of Bayesian models is the ability to test different kinds of prior knowledge and representations that people may bring to the problem at hand (e.g., rules, programs, or vector spaces \cite{Griffiths2010a}); in contrast, traditional neural networks lack an analogous mechanism for examining qualitatively different kinds of representations. Another advantage of Bayesian models, stemming from their use of richer priors, is their human-like ability to learn from just one or a few examples \cite{Tenenbaum1999,Tenenbaum2011,LakeScience2015}; in contrast, traditional neural networks are notoriously data hungry and can struggle to make meaningful inferences from just a few examples \cite{Geman1992}, although this characterization is changing with recent developments in architecture and pre-training \cite{Webb2024,WhitherSymbols}.

There are also challenges for Bayesian models that correspond to relative advantages of neural networks. Bayesian models risk oversimplification, both in how hypotheses are represented (e.g., the symbolic rules, grammars, or programs they use may only be first approximations) and through the parametric and independence assumptions used to define priors (e.g., Gaussian, categorical, or uniform distributions), limiting what they can capture about real-world structure. In contrast, neural networks typically rely on more general assumptions and can more flexibly capture the complexity of naturalistic data. (For two case studies, compare the Bayesian models for discovering semantic structure \cite{Kemp2008} or generating handwritten characters \cite{LakeScience2015} with related neural network approaches in \cite{RogersMcClelland2004} and \cite{Feinman2020a}, respectively.) The two approaches also differ in the computational demands of inference: making predictions with many Bayesian models requires averaging over large hypothesis spaces (Eq. \ref{eq_pp}), which can be intractable or require coarse approximations \cite{MacKay2003}; in contrast, neural networks are demanding to train, but once trained, making predictions requires only a fast forward pass.

\begin{figure}
\centering
\includegraphics[width=\linewidth]{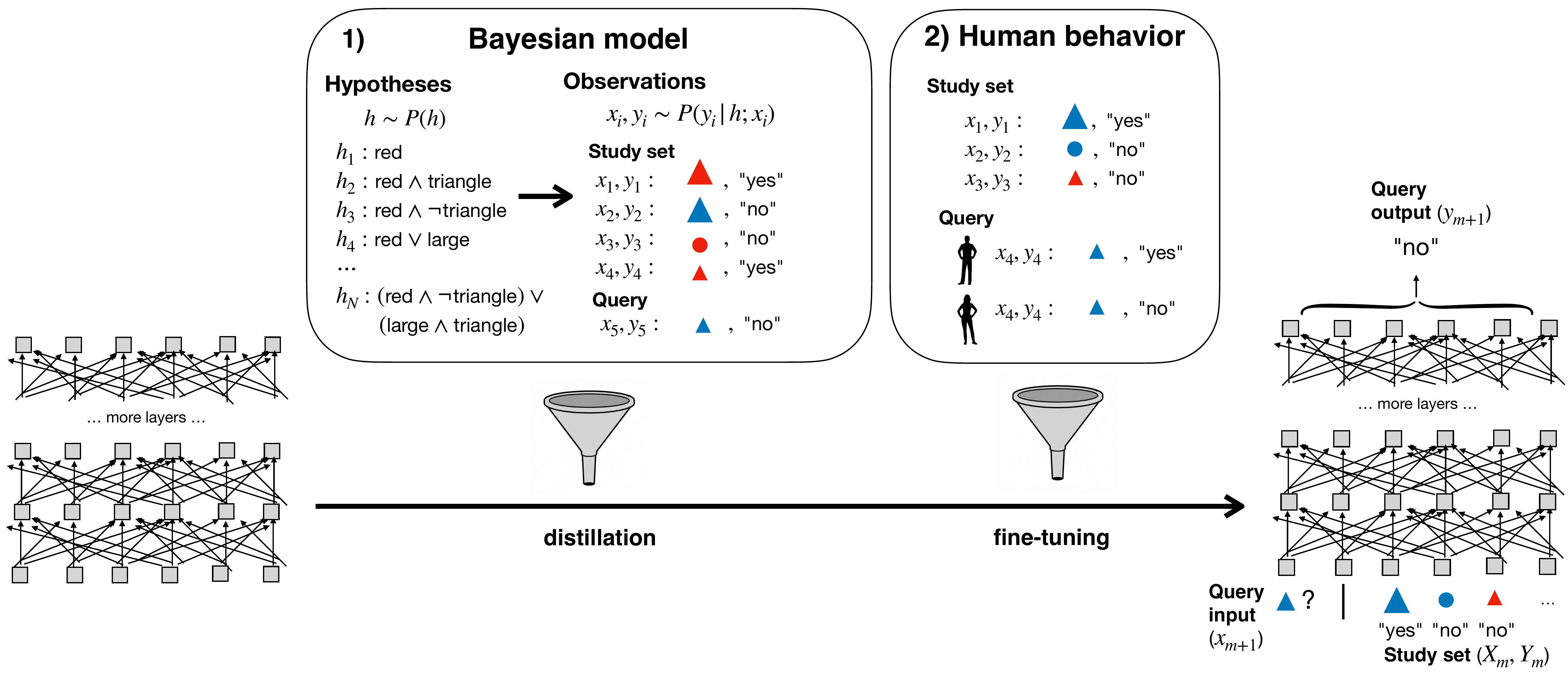}
\caption{The recipe for BBT modeling involves 1) distilling a Bayesian model into a neural network by training on synthetic data from forward sampling (no Bayesian inference required), and then 2) fine-tuning the network on human behavioral data (potentially very limited). During a training ``episode'', the network receives a query input $x_{m+1}$ (here, on the right, the network is queried on a small blue triangle) and a study set (objects $X_m$ and labels $Y_m$), which are concatenated and presented together as input. The network then produces a predicted query output $y_{m+1}$ (e.g., ``no'') which is compared to a target.}
\label{fig_bbt_method}
\end{figure}

Here, we introduce the Bayesian distillation with Behavioral Tuning (BBT) approach for achieving key advantages of both Bayesian and neural network models of human behavior and for building cognitive models that, like cognition itself, do not fall neatly into traditional categories, such as symbolic versus sub-symbolic representations or strong inductive biases versus generic learning processes (Fig. \ref{fig_bbt_intro}). Recent work \cite{McCoy2023} has proposed meta-learning as a means of bridging these traditions and Marr's levels \cite{Marr1982} more generally, allowing neural networks to acquire properties of Bayesian models through synthetic data \cite{Muller2022,LakeMLC,McCoy2023,jagadish2024human, jagadish2025meta,irie2025overcoming}. BBT builds on this work by introducing a modeling recipe for training powerful, task-specific transformer \cite{Vaswani2017} neural networks from scratch to predict and analyze human behavior, even when the behavioral data is highly limited. The two-stage procedure for building BBT models (shown in Fig. \ref{fig_bbt_method}) involves 1) specifying the desired prior knowledge and inductive biases as a Bayesian model and distilling them into a neural network through synthetic data, and 2) then fine-tuning that network on human behavioral data to capture additional predictable structure. This recipe allows BBT models to interpolate between modeling paradigms, as illustrated in Fig. \ref{fig_bbt_intro}: the first stage initializes the model as an approximation to a Bayesian model, and the second stage fine-tunes it away from that initialization in a data-driven fashion, adaptively stopping based on the quality of the predictions. We applied BBT to four diverse concept learning domains, including Tenenbaum's Number Game \cite{Tenenbaum1999}, Piantadosi, Tenenbaum, and Goodman's complex logical concepts \cite{Piantadosi2016c}, Shepard, Hovland, and Jenkins' classic boolean concepts \cite{Shepard1961}, and Lake and Baroni's compositional instruction learning \cite{LakeMLC}. Across each of these domains, the resulting BBT models can, like people, express strong prior knowledge, make computationally efficient inferences from just one or a few examples, and demonstrate complex behavior that goes beyond traditional Bayesian modeling assumptions.

A limitation is that although BBT aims to faithfully model human inductive biases, the two-stage model building procedure --- which incorporates the traditional strengths of Bayesian models and neural networks while also going beyond them --- does not reflect the process by which humans acquire those inductive biases.  We conclude with a discussion of how human minds could develop this distinctive set of computational strengths drawn from the two traditional approaches.

\section*{Modeling paradigm}
In this article, we focus on a family of learning tasks defined by a mapping $h \in H$ from inputs $x_i$ to outputs $y_i$. For instance, $h$ could be a logical rule, set of rules, or symbolic program. A learner observes a set of input examples $X_m = x_1,\dots,x_m$ (e.g., objects) matched to their corresponding outputs $Y_m = y_1,\dots,y_m$ (e.g., category labels), as specified by $y_i = h(x_i)$. Given a set of input-output examples for a given task ($X_m$ and $Y_m$), a BBT model seeks to predict how a human participant would respond ($\yprime$) to a new input ($\xprime$) (see Fig. \ref{fig_bbt_method} right).

The process of developing a BBT model is shown in Fig. \ref{fig_bbt_method}.  Developing a BBT model begins with a Bayesian model. The modeler uses the well-established Bayesian toolkit to specify prior knowledge about the task, or inductive biases they posit humans use to solve the problem. The prior $P(h)$ defines a hypothesis space $h \in H$ and how certain hypotheses should be favored over others. The likelihood $P(Y_m|h;X_m) = \prod_{i=1}^m P(y_i|h;x_i)$ specifies how the hypotheses $h$ relate to the input-output transformation, e.g., if $h$ is a logical rule as in Fig. \ref{fig_bbt_method} panel 1, then $P(y_i = 1 | h; x_i)$ could equal $1$ when $x_i$ follows the rule $h$ and $0$ otherwise. It follows through Bayes' rule that the posterior probability of a hypothesis given the current input-output mappings is $P(h|Y_m;X_m) \propto P(Y_m|h;X_m)P(h)$. Finally, relating to what BBT ultimately seeks to predict, Bayesian predictions for the output $\yprime$ in response to new input $\xprime$ can be computed as follows (see queries in Fig. \ref{fig_bbt_method} for examples),
\begin{align}
P(\yprime | Y_m; \xprime,X_m) & = \sum_h P(\yprime|h; \xprime) P(h|Y_m;X_m) \label{eq_pp} \\
& \approx f_\theta(\yprime ; \xprime, X_m, Y_m) \label{eq_pp_approx},
\end{align}
a quantity known as the posterior predictive distribution. Computing this quantity exactly requires a sum over all possible hypotheses $h$ that is generally intractable (Eq. \ref{eq_pp}). However, as described in the following paragraphs, BBT involves training a neural network ($f_\theta(\cdot)$; Eq. \ref{eq_pp_approx}) to approximate Eq. \ref{eq_pp} with just a forward pass; importantly, this approximation means that the sum in Eq. \ref{eq_pp} does not need to be computed. 
% Moreover, because BBT involves further fine-tuning the network $f_\theta(\cdot)$, the choice of hypothesis space $H$, prior, and likelihood (including their parametric forms and particular values) are not as delicate as in traditional Bayesian modeling of behavior. 

Given a Bayesian model, training a BBT neural network follows a two-stage process (Fig. \ref{fig_bbt_method}). In stage 1 (henceforth ``distillation''), the network $f_\theta(\yprime ; \xprime, X_m,Y_m)$ with parameters $\theta$ is pre-trained to approximate the posterior predictive distribution (Eq. \ref{eq_pp_approx}; see work on prior-data fitted networks \cite{Muller2022}). For the purposes of this article, $f_\theta(\cdot)$ is a transformer (although other choices are possible) that takes input $\xprime,X_m,Y_m$ (concatenated together and presented simultaneously) and produces a prediction about the output $\yprime$. For successful distillation, this network needs episodes sampled from the joint distribution $\yprime,\xprime,X_m,Y_m$ for pre-training (an example episode consists of the study set and query in Fig. \ref{fig_bbt_method} panel 1). Fortunately, it is straightforward to generate this kind of synthetic data from the Bayesian model by sampling $h$ from the prior and then sampling input-output pairs $\yprime,\xprime,Y_m,X_m$ based on the likelihood. Although producing these forward samples requires specifying the hypothesis space and the corresponding probabilistic model, it does not require posterior probabilities or posterior sampling. Pre-training the network requires many of these sampled episodes (indexed by $i$), each of which can be viewed as a mini-task defined by an unobserved $h^{(i)}$ and a mini-dataset based on observations $X_m^{(i)}$ and $Y_m^{(i)}$. Because the network is learning through gradient descent how to learn new episodes $i$ in-context, this can be viewed as memory-based meta-learning \cite{Hospedales2022,Muller2022,LakeMLC,binz2024meta}. At test time, all weights are frozen, and thus test-time learning of a new episode is implemented through just the forward propagation of activation patterns, which is distinct from in-weights meta-learning that instead adapts the weights to each new episode through gradient descent \cite{Finn2017a,McCoy2023}. If distillation is fully successful, the neural network would mimic the behavior of the Bayesian model.

The pre-trained network is not the finished product: BBT aims to make more accurate predictions than is possible with standard Bayesian modeling. To achieve this, in stage 2 (henceforth ``fine-tuning''; Fig \ref{fig_bbt_method}), the modeler fine-tunes the network $f_\theta(\cdot)$ from stage 1 on examples of human behavior. Specifically, the network is updated based on examples of participants responding with an answer $\yprime'$ to a query $\xprime'$ after observing $X_m,Y_m$.\footnote{One could consider fine-tuning the Bayesian model directly rather than using the proxy network $f_\theta(\cdot)$. However, this is often infeasible and/or undesirable: the Bayesian quantity Eq. \ref{eq_pp} is usually intractable, the neural network is differentiable while the Bayesian model is usually not, and a key purpose of BBT is not to rigidly follow a Bayesian specification.} 
Although pre-training helps to shape the fine-tuned network, behavioral fine-tuning is not constrained by the original prior $P(h)$, the original likelihood $P(Y_m|h;X_m)$, or even the logic of Bayesian inference. If supported by the human behavior, BBT could change the prior over hypotheses, how hypotheses relate to data, or add new hypotheses including those that defy straightforward symbolic description.

BBT was implemented by training the same 30 million parameter architecture for all four of the case studies, highlighting the generality of the approach, although some of the training hyperparameters were allowed to vary, as described in Section \ref{methods_arch}. The source code is available (\url{https://github.com/brendenlake/bbt-public}).

\section*{Results}
We applied BBT to four case studies focused on human concept learning, including number concepts, logical concepts, boolean concepts, and compositional functions.

\begin{figure}
\centering
\vspace{-2cm}
\includegraphics[width=\linewidth,height=0.87\textheight,keepaspectratio]{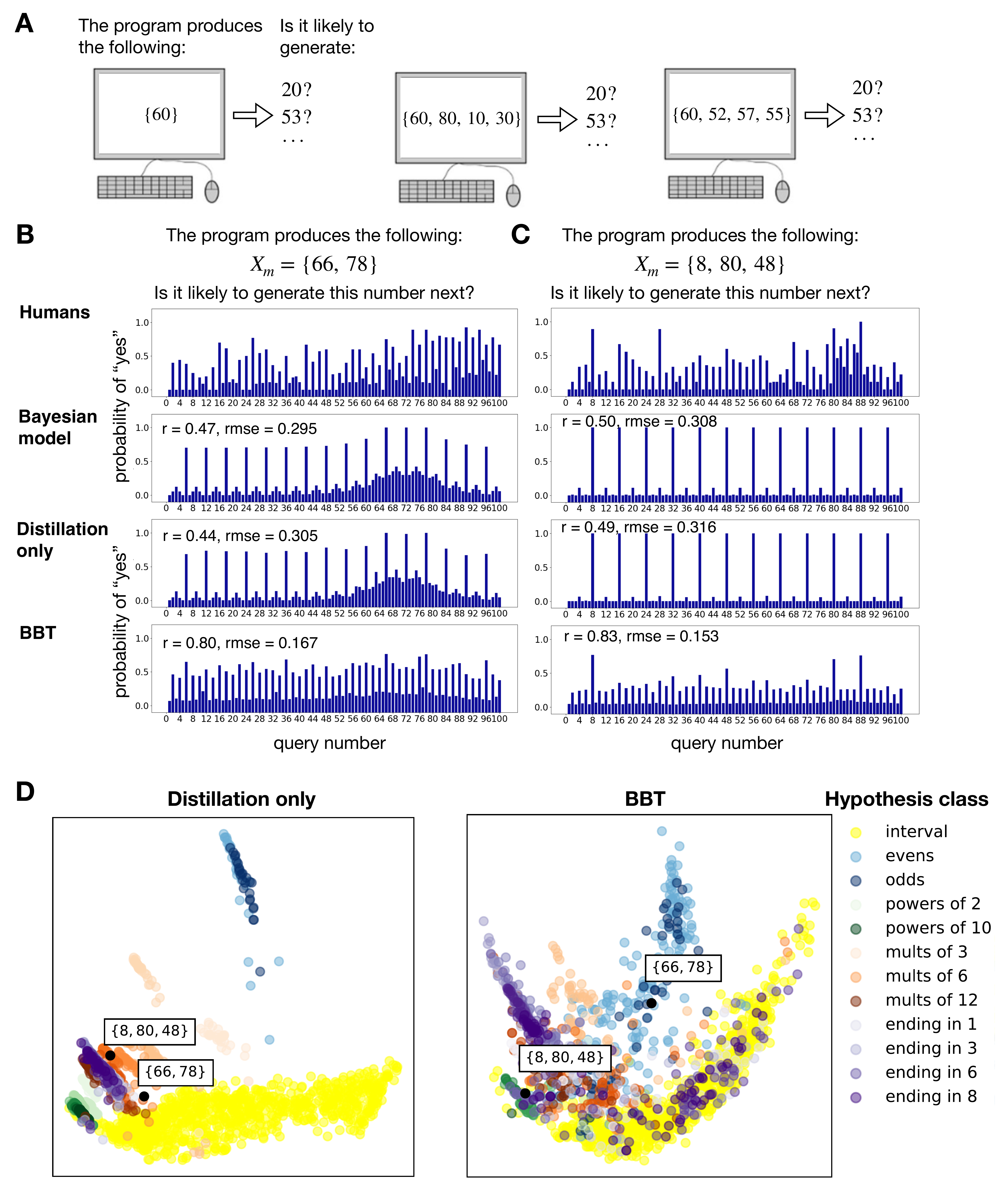}
\caption{The number game task and model predictions. A) Given a set of numbers that a mystery computer program produces (e.g., $X_m = \{60\}$), the task is to predict which other numbers the program is likely to generate (e.g., 20? 53?). The main text describes the three scenarios and how one might reason about them. B) Results for the study set $X_m = \{66,78\}$. The y-axis shows the proportion of ``yes'' answers for each query 1 to 100 (is it likely to generate this number next?) for humans and models. Also shown are the Pearson correlations ($r$) and root-mean-squared-error (RMSE) between the human and model judgments. C) Results for the study set $X_m = \{8,80,48\}$. D) PCA of embeddings after distillation only (left) and full BBT training (right), extracted from decoder layer 2. Each point in the scatter corresponds to a support set $X_m$ (averaged over all queries), marked with the color of its most likely hypothesis according to the fit sparse approximation. After fine-tuning with BBT, the influence of ``evens''/``odds'' (blue) and ``ending in n'' (purple) hypotheses dramatically expands.}
\label{fig_ngame}
\end{figure}

\textbf{Number concept learning.} The first case study is Tenenbaum's ``number game'' \cite{Tenenbaum1999,Tenenbaum2000}, a celebrated example of Bayesian modeling that is frequently used when teaching about the modeling paradigm. In the number game, an unknown computer program generates certain numbers from 1 to 100. Participants are provided with a small set of numbers generated by the program. As illustrated in Fig. \ref{fig_ngame}A (left), if all that is known so far is that the program produces $X_m=\{60\}$, then it's hard to make any confident predictions, e.g., whether the program is likely to generate 20, 53, or any other number.
However, if the program produces $X_m=\{60, 80, 10, 30\}$ (Fig. \ref{fig_ngame}A middle), then it seems likely to also generate 20 but not 53, based on a likely underlying rule that produces multiples of 10. 
If instead the program produces $X_m=\{60, 52, 57, 55\}$ (Fig. \ref{fig_ngame}A right), then it seems likely to also generate 53 but not 20, based on a likely underlying interval of numbers. The goal of the modeling is to predict human behavioral judgments. For this purpose, we used a human dataset from Bigelow and Piantadosi of yes/no generalization judgments on the basis of 255 unique study sets ($X_m$) of numbers that the program produces \cite{Bigelow2016}. 

We fit BBT to study what Bayesian models of the number game could be missing. We used Tenenbaum's Bayesian model for distillation \cite{Tenenbaum1999,Tenenbaum2000}. In this model, each hypothesis $h \in H$ is a set of numbers allowed by the hypothesis. The space $H$ includes mathematical hypotheses (evens, odds, multiples of $n$, powers of $n$, etc.) and interval hypotheses (all numbers within a certain range), for a total of 5084 hypotheses. The prior $P(h)$ specifies the extent to which mathematical or interval hypotheses are favored, and the extent to which medium-sized intervals are favored. Assuming that a number $x_i \in h$ is sampled from uniformly from all numbers in  $h$, then the likelihood is $P(x_i|h) = \mathbbm{1}\{x_i \in h\}\frac{1}{|h|}$, such that $\mathbbm{1}\{x_i \in h\}$ is 1 or 0 based on membership. (We drop $Y_m$ for convenience because for all $x_i \in X_m$, we define $y_i = 1\{x_i \in h\} = 1$. Section \ref{methods_bayes_num_game} of the Methods has more details about the change in notation for this case study.) After pre-training on 100,000 synthetic episodes, BBT is fine-tuned on behavioral examples based on 178 number sets, with 26 sets for validation and 51 held-out sets for final testing of the model predictions. A control model (``distillation only''; stage 1 but not stage 2 training) has the same architecture as BBT and was pre-trained exactly like BBT was, although it was not fine-tuned on human behavioral data. Another control model (``fine-tuning only''; stage 2 but not stage 1 training) was fine-tuned on behavioral data but not did not benefit from the Bayesian prior through distillation. Note that, by skipping the Bayesian distillation, this model is most aligned with traditional neural network training.

A preliminary question is how well this Bayesian model can be distilled into the transformer. For a set of numbers $X_m$ that the program generates, the models estimate whether a query $\xprime$ is also endorsed as likely, $P(\xprime \in h | X_m)$. As shown in Fig. \ref{fig_ngame}B and C, the Bayesian model and the pre-trained network typically make very similar predictions. Indeed, on the held-out sets, their predicted probabilities of endorsing queries are strongly correlated (Pearson $r=0.997$, $n=5100$, aggregating across all 51 held-out sets).

The next question is which modeling approach best predicts human behavior. First, as summarized in Table \ref{table_ll}, the models were evaluated based on the log-likelihood of the held-out human responses. BBT provides the best predictions by a margin of 3,175 log points, followed by the original Bayesian model and the distillation only model. The fine-tuning only model has the lowest performance, highlighting the need for distillation in this case. Second, the models were evaluated by computing the Pearson correlation coefficient and root-mean-squard-error (RMSE) between the model's probability of endorsing a query with ``yes'' and the proportion of humans that responded with ``yes'', aggregated across all query $\xprime \in h$ and support $X_n$ combinations. BBT provides the best fit using either metric ($r=0.81$, $\text{RMSE}=0.153$, $n=5100$), outperforming the Bayesian model ($r=0.66$, $\text{RMSE}=0.302$), the distillation only network ($r=0.65$, $\text{RMSE}=0.307$), and the fine-tuning only network ($r=0.54$, $\text{RMSE}=0.221$).

To better understand the source of BBT's gains, we also examined whether the original Bayesian model, despite its parametric form, can be fine-tuned like a neural network to better predict the human behavior. The performance of this model compared to BBT can help measure the degree to which BBT's gains come from fine-tuning more generally, versus fine-tuning neural networks more specifically, with more flexible assumptions. To examine this, we initialized the Bayesian model's parameters with their original values \cite{Tenenbaum1999}, and because this particular model's form happens to be tractable and interpretable as a multilayer perceptron (see Extended Data Fig. \ref{fig_ngame_alt} for illustration and optimization details; note that this is a special case that does not apply to most Bayesian models), we could fine-tune the Bayesian model on the same training set of 178 prompts. We compared two variants with varying sets of learnable parameters, including a prior-fit model with a different probability $P(h)$ for all 5,084 hypotheses (5,085 parameters including lapse rate; also explored in \cite{Bigelow2016}) and a full-fit model with a different prior probability $P(h)$ and a fully-flexible likelihood for each hypothesis $P(x_i|h)$ (518,569 parameters; the likelihood of each hypothesis is an arbitrary categorical distribution).\footnote{Informally, including an additional parameter for posterior tempering did not substantially change the results.} Neither the prior-fit Bayesian model (-31,419.3 log-likelihood, $r=0.67$, $\text{RMSE}=0.301$; compare with Table \ref{table_ll} column 1) nor the full-fit Bayesian model (-28,736.7, $r=0.78$, $\text{RMSE}=0.166$) achieved the level of performance that BBT does (the latter separated by 578.8 natural log points), consistent with the motivation that BBT's relaxation of parametric assumptions allows it to better predict the human behavior.

\begin{table}
\centering
\resizebox{\textwidth}{!}{%
\begin{tabular}{l||llll}
Model                     & Number concepts & Logical concepts & Shepard concepts & Compositional rules \\
\hline
Bayesian          & -31,332.6 & -79,023.3 & -5,554.4 &  \\
Distillation only & -31,435.7 & -77,249.9 & -5,601.1  & -466.4\\
Fine-tuning only  & -31,465.2 & -106,926.3 & -5,924.7 & -1,653.6\\
BBT              & \textbf{-28,157.9} & \textbf{-69,423.8} & \textbf{-5,260.8} & \textbf{-343.6}
\end{tabular}
}
\caption{Model predictions evaluated through the log-likelihood of held-out human behavior. The bold shows the best scoring model (higher is better). All models had lapse rates that were fit, as described in Section \ref{methods_fitting} of the Methods. Neural networks (Distillation only, Fine-tuning only, and BBT) were trained for five runs, with the reported run selected through a separate validation set.} \label{table_ll}
\end{table}

% previously, $X_n = \{66, 78\}$ and $X_n = \{31, 3, 1, 15\}$
What kinds of structure is BBT capturing that the Bayesian model is not? The held-out sets $X_m = \{66,78\}$ and $X_m = \{8,80,48\}$ in Fig. \ref{fig_ngame} were further analyzed using a sparse decomposition of BBT's predictions (see Section \ref{methods_sparse} of the Methods for details), as shown in Extended Data Figs. \ref{fig_ngame_66_78} and \ref{fig_ngame_8_80_48} respectively, visualizing the hypotheses with the strongest influence in BBT's predictions. For the program that produces $\{66,78\}$, the Bayesian model narrows in on the hypothesis ``multiples of 6'' (with posterior probability 0.53) that people (and BBT) largely miss or find less compelling. In contrast, BBT's predictions are consistent with ``evens'' at probability 0.46 while the Bayesian model has it at 0.05. BBT predictions for this support set are also consistent with ``ending in 6'' and ``multiples of 6'', while the Bayesian model does not consider the former because the hypothesis includes 66 but not 78, suggesting a substantial change in the BBT representation compared to the original Bayesian model. For the program that produces $\{8,80,48\}$, the Bayesian model narrows in on ``multiples of 8'' with probability 0.89. In contrast, BBT's predictions (and the human predictions) are also consistent with ``ending in 8'' and ``evens.'' Note, however, BBT is not a perfect match with the human data on this episode: people also endorsed all the numbers in the 80s, which BBT does not predict.
%% Figure out why human judgementns are not at ceiling, even for provided numbers

The difference between the Bayesian model (as mimicked in ``distillation only'' training) and BBT is also evident in their internal representations. Fig \ref{fig_ngame}D shows how the networks learn to organize various support sets ($X_m$) according to their likely hypothesis class (displayed as distinct colors; see Section \ref{methods_sparse_recovery} of the Methods for the specifications of this simulation), despite never seeing these labels during training. The vector embeddings for varying support sets were extracted after the second layer of the transformer decoder (other layers can be used too), and then principal component analysis (PCA) was applied to visualize the embeddings in two dimensions, accounting for 98.7\% and 90.0\% of the variance for the distillation only and BBT models, respectively. The first principal component (x-axis) is strongly correlated with the probability of answering ``yes'' ($r=0.92$ for distillation only and $r=0.81$ for BBT), and together with the second component, leads to evident clusters (listed left-to-right) for ``powers of n'' (greens; light to dark with increasing n), ``ending in n'' (purples), ``multiples of n'' (reds), ``evens'' (light blue), and then ``odds'' (dark blue). After BBT fine-tuning, the ``evens'', ``odds.'' and ``ending in n'' hypothesis classes have much greater influence, taking over and reorganizing parts of space previously dominated by other classes (especially intervals and powers), a shift that is quantified in Extended Data Fig. \ref{fig_complexity_diff}A. For instance, rather than residing in the ``multiples'' classes, the embedding for $\{66,78\}$ is now in part of the space dominated by ``evens'' (consistent with the above behavioral analysis), and $\{8, 80, 48\}$ is now lumped with other classes including ``ending in n'' (also consistent with the behavior).
%% Include "output: fig_ngame_category_diff.pdf" in the supporting figures

\begin{figure}
\centering
\includegraphics[width=\linewidth,height=0.72\textheight,keepaspectratio]{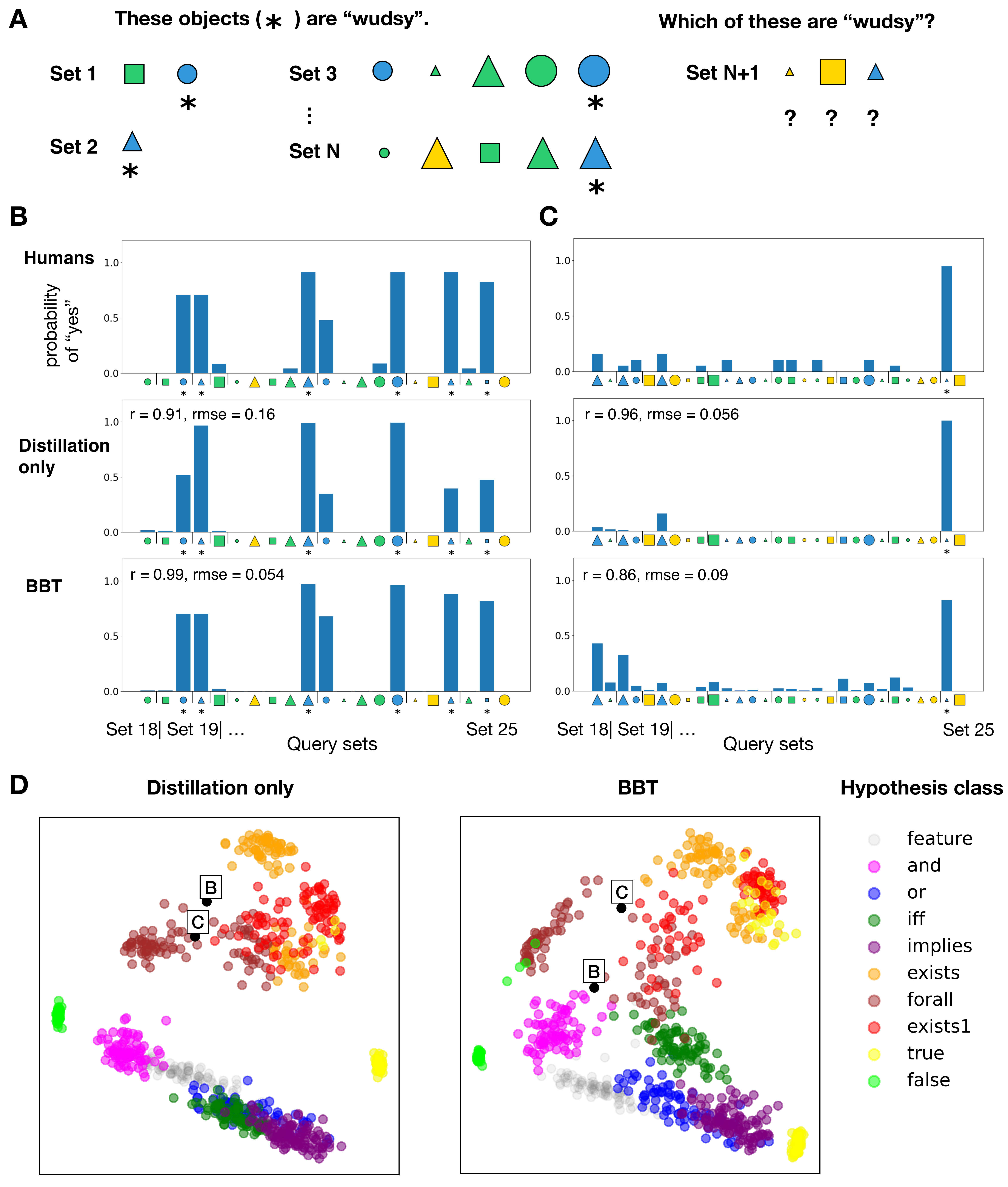}
\caption{The logical concept learning task and model predictions. A) Given $N$ sets of objects and their category labels (positive examples marked with $*$), the task is to predict the labels for the $(N+1)$th set. Here, the correct rule is ``largest blue object in the set.'' B) Human and model judgments on the final 8 object sets when learning the rule in Panel A. The y-axis shows the proportion of ``yes'' answers per object. The x-axis shows the object stimuli, with the object sets divided by vertical lines. Also shown are the Pearson correlations ($r$) and root-mean-squared-error (RMSE) between human and model judgments. C) Judgments on the last 8 sets when learning the rule ``is the only blue or green object in the set.'' D) PCA of model embeddings after the distillation (left) and human fine-tuning (right) stages of BBT training, extracted from decoder layer 3. Each point is a different learning episode containing 24 object sets as study (averaging over 20 different query sets), marked with the color of its most likely hypothesis as fit to the model judgments via sparse approximation. After BBT fine-tuning, the model blurs the separation between Boolean vs. non-Boolean episodes, and the simplest hypotheses (true, false, individual features) expand their influence.}
\label{fig_set_concepts}
\end{figure}

\textbf{Logical concept learning.}
The second case study is Piantadosi, Tenenbaum, and Goodman's large-scale examination of logical concept learning \cite{Piantadosi2016c}. Studies of how people learn logical rules have a long history \cite{Shepard1961,Bruner1956} and feature prominently in investigations of whether human category representations are best understood as  symbolic \cite{Goodman2008a}, sub-symbolic \cite{Kruschke1992}, or hybrid \cite{Erickson1998}. Piantadosi et al.'s study is distinctive in the scope and sophistication of the logical concepts that people were asked to learn. As illustrated in Fig. \ref{fig_set_concepts}A, participants were presented with a set of objects (ranging in size from 1 to 5) and  asked to categorize each object as ``wudsy'' or not. Participants were then given feedback and asked to make judgments about a new set while all the previous sets remained visible on the screen, for a total of 25 sets. Of the 108 possible concepts, 34 were defined by rules in Boolean (propositional) logic, consisting of only features (color, shape, size) and Boolean operators (and, or, implies, not), e.g., an object is ``wudsy'' if it is ``blue or a circle''. For these rules, each object's membership was meant to be decided by considering that object in isolation. For more complex (non-Boolean) concepts, each object's membership was meant to be decided in relation to the other objects in the corresponding set as specified through quantifiers, e.g., an object is ``wudsy'' if it is ``the largest blue object in the set.'' Averaged across all concepts, people were able to achieve 78\% correct, which can be compared to a baseline of always answering ``no'' which achieves 56\% correct. Considering that people were well-above chance but far from perfect, modeling human learning as Bayesian inference over logical rules may not be the complete story.

We fit BBT to examine whether hybrid Bayesian-connectionist modeling could better predict human behavior. As is necessary for building BBT models, we used Piantadosi et al.'s \cite{Piantadosi2016c} Bayesian models to bootstrap the BBT modeling. In their prior work, the authors \cite{Piantadosi2016c} compared 21 different Bayesian models to the human behavior, varying in the operations they allow to appear in hypotheses $h \in H$, which are rules that evaluate to True or False for a given object. We take their most successful model as the starting point. For this model, the rules $h \in H$ can reference object features (circle, triangle, rectangle, yellow, green, blue, and sizes 1 to 3), boolean operators (described previously), relations that compare objects through their features (same shape, same color, etc.), and quantifiers (``for all'', ``exists'', and ``exists one or fewer''). The prior $P(h)$ is a probabilistic context-free grammar that favors shorter rules; although the original model \cite{Piantadosi2016c} fit the production probabilities of the grammar, we used uniform probabilities for additional simplicity (except two production probabilities were upweighted five-fold to control the complexity of the expressions: the weight for sampling a non-trivial expression, and the weight for sampling a feature rather than a boolean operator, relation, or quantifier). For a set of $k$ objects $x_i$ and their labels $y_i \in \{0,1\}^k$, the likelihood $P(y_i|h;x_i)$ is equal to 1 if all the labels are consistent with the rule, and 0 otherwise. (The original model \cite{Piantadosi2016c} had high levels of noise in the likelihood, which we removed here.) After distillation of the Bayesian model on a set of 200,000 synthetic episodes, BBT was fine-tuned on the human behavior. Piantadosi et al. \cite{Piantadosi2016c} provided a training/test split we used for this: each of the 108 concepts had a learning sequence of 25 sets for fine-tuning and another learning sequence of 25 sets for held-out testing.

The models were evaluated for goodness-of-fit based on the log-likelihood of the human behavior from the held-out learning sequences (Eq. \ref{eq_pp}). A summary of the results is shown in Table \ref{table_ll} column 2. BBT is the best fitting model by a margin of at least 7,826 natural log points. BBT makes more accurate behavioral predictions than the 21 Bayesian models considered in \cite{Piantadosi2016c}; the best of these models is listed in Table \ref{table_ll}. BBT also outperforms the distillation only network and the fine-tuning only network by large margins. Considering the additional metrics of correlation and RMSE, aggregating across all the test queries for the final 8 sets per task, BBT also provides the best fit (BBT $r=0.88$, $\text{RMSE}=0.190$, $n=2,657$; Distillation only $r=0.87$, $\text{RMSE}=0.214$; Fine-tuning only $r=0.57$, $\text{RMSE}=0.325$; the final 8 sets for two example concepts are shown in Fig. \ref{fig_set_concepts}B and C).

We further examined BBT's generalization capabilities to new concepts because Piantadosi et al.'s train/test split was not designed for powerful data-driven models. Specifically, the held-out learning sequences, while novel, were consistent with the same logical concepts that were used for BBT fine-tuning. Thus, to ensure BBT can generalize to new concepts, we zoomed in on the model performance for just the concepts in the test set that did not appear in the learning sequences for fine-tuning (that is, the 11 concepts shared with the validation set). Across all five pre-training runs, BBT fine-tuning improves the performance of the pre-trained network by at least 481.7 log points on held-out concepts (see Extended Data Table \ref{table_ll_logical_novel}), consistent with generalization to novel concepts.

What additional structure does BBT learn during the fine-tuning process on human behavior? The distinction between more traditional Bayesian accounts (including our distillation only network) and BBT is not simply a matter of accuracy in predicting the ground-truth rule, as both models are equally accurate in our case (69\% correct). This stems from the fact that Piantadosi et al. \cite{Piantadosi2016c} asked people to learn concepts that were either highly unlikely or outside the Bayesian hypothesis space; in our case, a rule consistent with the ``largest blue object'' task in Fig. \ref{fig_set_concepts}B was never sampled during distillation. Nevertheless, even in such cases where the distillation only network diverges from the ground-truth rule, the distillation only network can make meaningful and subtle predictions, and so can the resulting BBT model after fine-tuning. For instance, in that same task shown in Fig. \ref{fig_set_concepts}B, people and both models partially endorsed the first blue object in set 23 (Fig. \ref{fig_set_concepts}B), even though it is not the largest in the set, suggesting that the simpler ``blue'' hypothesis is an appealing alternative (especially for the BBT model, as shown in a sparse approximation analysis; Extended Data Fig. \ref{fig_set_is_largest_blue}). Another observation is that fewer people endorsed the blue objects in the smallest sets (sets 19 and 20), seemingly reflecting uncertainty about how the quantifier applies to small sets, an effect that BBT captures better than the distillation only network. In another example, when tasked with learning ``the only blue or green object in the set'' (Fig. \ref{fig_set_concepts}C), some people and BBT were tempted by blue or green objects more generally without the quantifier constraints, also consistent with a greater emphasis on simpler hypotheses in BBT compared to the distillation only network (Extended Data Fig. \ref{fig_set_is_only_blue_green}).

These findings are echoed in a PCA analysis of the internal representations (Fig. \ref{fig_set_concepts}D; see Section \ref{methods_sparse_recovery} of the Methods for the specification of this simulation). The embedding space of the distillation only network has clear organizational structure, largely separating the different classes of hypotheses. Notably, the second principal component (y-axis) separates the simpler Boolean hypotheses (bottom) from the more complex non-Boolean hypotheses that use quantifiers (top), which is a key organizing principle for the candidate concepts. However, the models' organization is partially  driven by lower-level correlates; we found that the x-axis correlates $r=0.974$ with the first output probability of ``yes'', and the y-axis correlates $r=0.874$ with the binary entropy of that probability (relating to model ``confidence''), although these components explain just 54.4\% of the total variance in the embeddings. After fine-tuning on human behavior, the resulting BBT model maintains much of this internal structure, including the above two correlations at $r=0.986$ and $r=0.877$, respectively. Notably, however, the separation between the Boolean and non-Boolean hypotheses narrows and blurs, which mirrors the human and model behavior in Fig. \ref{fig_set_concepts}. Additionally, the simplest hypotheses expand in their influence in the BBT embedding space (true, false, and individual features) at the expense of the hypotheses with quantifiers, a shift we quantified using the sparse approximation analysis in Extended Data Fig. \ref{fig_complexity_diff}B-i. This analysis confirms a shift from more complex to simpler hypotheses: on average, the best-fitting hypotheses for BBT had 2.2 fewer function calls (Extended Data Fig. \ref{fig_complexity_diff}B-ii).

\textbf{Shepard concept learning.} The third case study is the concept learning task from Shepard et al. \cite{Shepard1961}, which has been reproduced by several groups \cite{nosofsky1994comparing, Lewandowsky2011, rehder2005eyetracking, kurtz2013human} and has been conducted on different populations \cite{Badham2017}. We chose this task because it allows us to examine how BBT can be used to model different populations and how BBT generalizes to behavior from replication studies collected years apart.

\begin{figure}
\centering
\includegraphics[width=\linewidth,height=0.68\textheight,keepaspectratio]{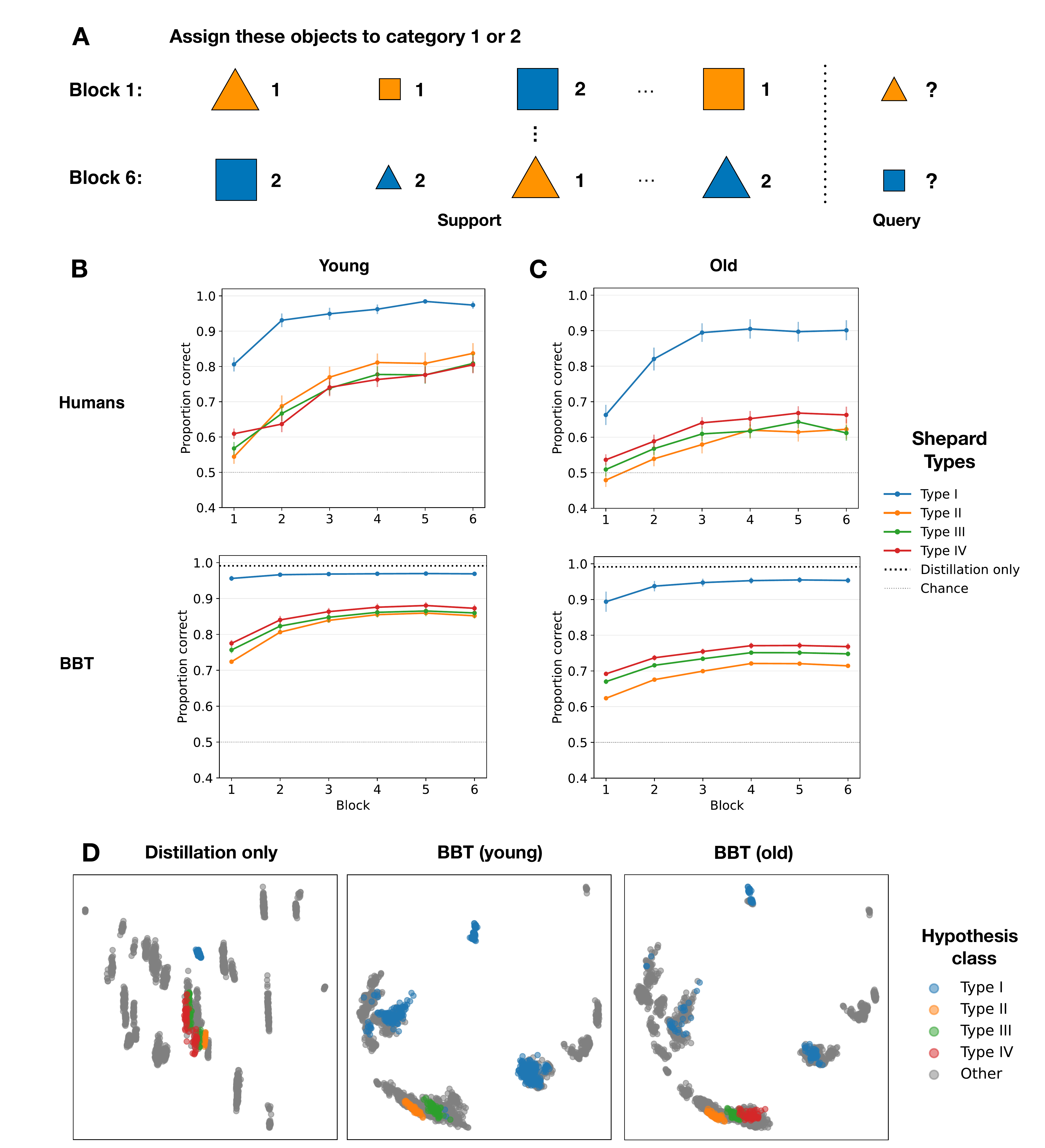}
\caption{The Shepard concept learning task and model predictions. 
A) Participants are presented $N$ objects sequentially, tasked with choosing between two categories (1 and 2) on each trial and provided with corrective feedback. The stimuli are repeated over multiple blocks. In the example shown, the correct rule is ``all yellow objects belong to category 1''. 
B \& C) Learning curves over blocks for humans, distillation only networks, and BBT for young (B) and older (C) participants, data taken from the Badham et al. \cite{Badham2017} study. The y-axis shows the proportion of correct choices and x-axis shows the block number, with the four Shepard category structures (Type-I to Type-IV) shown as distinct lines.  
D) PCA of model embeddings after the distillation (left), BBT (conditioned on young embedding; middle), and BBT (conditioned on older group embedding; right) extracted from decoder layer 3. Each point is a different learning episode containing 16 objects in support and all unique objects as query, marked with the color of its most likely hypothesis from four out of six Shepard's category learning structures as fit to the model judgments via sparse approximation.  The first component in BBT (old) can clearly separate Types II-IV (orange, green, and red) and Type I (blue), whereas in BBT (young) younger participants only separate Type II from Type III, and the second principal component separates Type~I (blue) from other types for both BBT conditioned on young and old. Although both BBT (young) and BBT (old) show stronger representation of Type I hypothesis compared to Distillation only, the increase is stronger in younger than older. Notably, no episode is best described by a Type IV hypothesis in BBT (young): all Type IV episodes are instead fit by simpler non-Shepard rules (2–3 literals), whereas BBT (old) retains most of them (53 of 60).
% D) Log-likelihood of human behavior from the held-out (test) participants in the Badham et al. study under the Distillation only, BBT, and Fine-tuning only model. The hatched bar indicate in-domain transfer, which how well fine-tuning on subset of participants in the Badham et al. \cite{Badham2017} study generalize to its own held-out participants,  and solid bar indicate cross-domain transfer, which is how well fine-tuning on subset of participants from a different Shepard replication study (specifically, Lewandowsky et al. \cite{Lewandowsky2011}) generalizes to held-out participants from the Badham et al. \cite{Badham2017} study. 
}
\label{fig_shepard}
\end{figure}

In Shepard's concept learning task, participants learned to classify objects that vary along three binary-valued features (shape, color, and size) into two categories. As shown in Fig. \ref{fig_shepard}A, objects were presented sequentially and participants predicted their category after each presentation. Unlike in the other case studies, this task imposes memory demands because the previous trials were not displayed concurrently to the participants.
Training consisted of up to six blocks of 16 trials, with each of the eight stimuli appearing twice per block in randomized order. Learning performance was measured until they perform perfectly in two consecutive blocks or reach the maximum number of trials, whichever comes first.
Shepard et al. considered six non-trivial, balanced category structures, corresponding to rules of increasing logical complexity \cite{Feldman2000}: Type~I is
a single-feature rule (as illustrated in Fig. \ref{fig_shepard}A), Type~II is a two-feature XOR, Types~III--V combine a one-feature rule with one or two exceptions, and Type~VI has no short logical regularity. A key finding is that the participants display the same difficulty ordering as predicted by the Boolean complexity of the underlying category structure \cite{Feldman2000}: Type $\mathrm{I}<\mathrm{II}\approx\mathrm{III}\approx\mathrm{IV}\approx\mathrm{V}<\mathrm{VI}$, a shown in the first row of Fig. \ref{fig_shepard}B-C.

To fit a BBT model, we started with a Bayesian model over symbolic hypotheses and distilled it into a neural network. We chose the Rational Rules model \cite{Goodman2008a}, which operationalizes category learning as Bayesian inference over rules in Boolean logic (that is, rules that use only conjunctions, disjunctions, and logical not; see Fig. \ref{fig_bbt_method} for an illustration). The prior $P(h)$ is specified using a probabilistic context-free grammar that favors shorter rules. The production probabilities of the grammar are sampled from a Dirichlet distribution, keeping the production rules fixed \cite{Goodman2008a}; see Section~\ref{methods_bayes_shepard} of the Methods for details of implementation.
% The likelihood $P(y_i \mid h; x_i) = 1 - \frac{\epsilon}{2}$ if $h$ is consistent with the observed label and $\frac{\epsilon}{2}$ otherwise, where $\epsilon = 0.01 $ (see Section~\ref{methods_bayes_shepard} of the Methods for details of implementation). 
After distillation on a set of 1,000,000 synthetic episodes, BBT was fine-tuned on human behavioral responses from Badham et al. \cite{Badham2017}. The study collected data from 48 young participants (age 18-21) and 48 older participants (age 60-87) on four (Types~I--IV) of the six categories structures over 6 blocks (x 16 trials per block = 96 trials). Critically, we learned an additional block-specific and age-specific embedding directly from the data during fine-tuning to capture learning over blocks and age-based differences (see Section~\ref{methods_bayes_shepard_2} of the Methods). 

We find that BBT reproduces both the learning trajectory across blocks and its interaction with age when conditioned at inference time on block (1-6) and age group (young or older) (Fig.~\ref{fig_shepard}B-C and Extended Figures \ref{fig_shc_extended_2}). Specifically, BBT showed strong correlations with the human learning trajectory  ($r=0.938$), unlike the distillation only network, which akin to Rational Rules \cite{Goodman2008a} only captures aggregate effects and thus the correlation is undefined, and the fine-tuning only model ($r=0.022$; see Extended Data Figure \ref{fig_shc_extended_2} last row). Furthermore, BBT shows the age-related accuracy gap observed in human participants although its overall accuracy is higher (BBT simulation: 86.9\% for younger adults and 77.7\% for older adults; participants: 78.6\% and 65.9\%, respectively).
% whereas the fine-tuning-only model collapses Types~II--IV together in the young condition and under predicts the age gap. only reproduces the ordering effect but does not display any block-based learning. 

% We considered two types of data split to model human behavior. First, 
We also studied BBT's ability to generalize to unseen participants, dividing the participants (or specific sub-group of participants, like young or older adults) into 65\%, 10\% and 25\% splits for training, validation, and testing, respectively. Concretely, we evaluated goodness-of-fit based on log-likelihood of human behavior from the held-out (test) participants; see column 3 in Table \ref{table_ll}.  
BBT is the best fitting model with log likelihood of -5,260.8 natural log points, outperforming Rational Rules  (-5,554.4),  the distillation only (-5,601.1) and the fine-tuning only (-5,924.7).

To better understand where BBT's gain comes from, we view it through the lens of sparse hypothesis decompositions (see Fig.~\ref{fig_shepard}D and Extended Figures \ref{fig_shc_extended_1}A). 
We applied PCA to the embeddings (at the end of block 1) extracted from the third layer of the decoder to visualize them in two dimensions. 
The two components together capture more than $85\%$ of the variance between models, with the first component strongly correlated with the probability of answering "yes" ($r = 0.99$ distillation only; $r = 0.92$ for BBT conditioned on young and $0.94$ for BBT on old), and the second component correlates with the entropy of the model's query predictions ($r = 0.74$ and $0.76$ for BBT young and old, respectively).
On visual inspection, we find that the first component in BBT (old) can clearly separate Types II-IV (orange, green, and red) and Type I (blue), whereas in BBT (young) younger participants only separate Type II from Type III. The second principal component separates Type~I (blue) from other types for both BBT conditioned on young and old; see Extended Data Figure \ref{fig_shc_extended_1}D for visualization based on the complexity of rules instead. 
% BBT (young) clearly shows a stronger representation of the Type~I hypothesis than BBT (old), compared to the Distillation only model. 
At an aggregate level, we find that the complexity of the rules used for categorization decreases in BBT compared to Distillation only, with the proportion of hypotheses representing Type I increases in BBT, more so in BBT in younger participants ($0.02\rightarrow0.17$) rather than older participants ($0.02\rightarrow0.05$), while for all other types it decreases; see Extended Data Figures \ref{fig_shc_extended_1}A and C. 
This suggests that at the end of block 1 participants (both young and old) tend to favor simpler Type~I rules that operate on single features beyond what is prescribed by the Rational Rules prior, with younger participant inferring it much more quickly than older participants.

Finally, we examine how well BBT fine-tuned on behavioral responses from one Shepard replication study \cite{Lewandowsky2011} generalizes to predict behavior in another replication \cite{Badham2017}, which was collected six years later.
This serves as a stronger test for generalization, in comparison to generalization to unseen participants in the same experiment, as it evaluates how effectively can BBT predict behavior collected by a different experiment that differs in presentation-format, task instruction, and population-sample. 
To do this, we evaluated how well the BBT model \emph{fine-tuned on the Lewandowsky study} can predict behavioral responses from held-out participants in the \emph{Badham study}
\footnote{The inverse was not considered since the Lewandowsky study (113 adult only participants) had more task blocks (12 instead of six) and category types (six instead of four) than the Badham et al. study.}. 
When evaluated on held-out participants from the Badham study, BBT (-5,331.1) fine-tuned on the Lewandowsky dataset achieves a log-likelihood close to the BBT fine-tuned on (train-split of) the Badham dataset (-5,298.7), and clearly exceeds fine-tuning only (-5,926.5) and Distillation only (-5,565.7). This demonstrates BBT's capacity to generalize between studies in the Shepard category learning task. 
Furthermore, as the distillation only network is better than both the fine-tuning only networks but still worse than the BBT models, it indicates that the rational rules prior distilled into BBT provides a substrate on which behavioral fine-tuning transfers across behavioral datasets.

% [PLACEHOLDER] Finally, we viewed model differences through the lens of sparse hypothesis
% decompositions to better understand where the BBT's gain comes from (Methods Section~\ref{methods_sparse};
% Fig.~\ref{fig_shepard}F).
%  Under Distillation only, PCA of the hypotheses most influential
% in model predictions captures the complexity of the underlying hypothesis along PC1,  
% Type~I from all other types along PC1
% (\textbf{66.8\%} variance), with Types~II--V collapsing into a single tight
% cluster --- the same collapse produced by the Bayesian prior. BBT preserves
% the Type~I separation but pulls Types~IV--VI apart, with Types~IV and V
% moving toward shorter, partial-rule hypotheses that match most-but-not-all
% exemplars (the near-rule representations people tend to fall back on when
% the fully-consistent rule is hard to find), while Type~VI moves in the
% opposite direction toward longer, memorization-like hypotheses. 
% Higher complexity for rules in the BBT model than for the Distillation only model.
% This
% parallels the number-game finding that BBT's fine-tuning relaxes the
% Bayesian prior's commitment to a narrow rule set and broadens the effective
% hypothesis space toward the partial rules humans actually use.

\textbf{Compositional instruction learning.} In the final case study, we fit BBT models to predict how people learn sets of compositional rules from examples. Unlike the previous case studies, which involve learning a single underlying rule for each task, the compositional instruction-learning task introduced by Lake and Baroni \cite{LakeMLC} involves learning multiple interacting rules that jointly explain the study (input-output) examples. Success on the subsequent queries depends on compositional generalization: applying learned rules to novel primitives and composing more rules than were observed during study. Such systematic generalization has long been viewed as a hallmark of human minds and a particular challenge for neural networks  \cite{Fodor1988,LakeBaroni2018}. Recent work shows how compositional skills can be acquired by neural networks through distillation \cite{LakeMLC}, providing a foundation for the current case study. However, this previous work relied on  hand-designed priors to reproduce the characteristic patterns of human behavior. Instead, here we show that the appropriate inductive biases can be acquired through BBT without the need for hand engineering.

\begin{figure}
\centering
\includegraphics[width=\linewidth]{figures/fig-miniscan.jpg}
\caption{Few-shot learning of compositional instructions. The task involves responding to strings of pseudowords (inputs) with sequences of colors (outputs). A) Participants learned these 14 study instructions which are composed of four primitives and three functions (heading were not provided to the participants). B) After learning the study instructions, participants were asked to respond to 10 query instructions (4 of 10 are shown here). The most frequent 4 responses to each query instruction are shown, labeled in parentheses with the count for people and the percent of samples for the models. Superscripts indicate which responses are correct ($^*$) or consistent with one-to-one biases (1-to-1) or iconic concatenation biases (IC).}
\label{fig_miniscan}
\end{figure}

The instruction learning task is shown in Fig. \ref{fig_miniscan}. Human participants were asked to learn a set of 14 study instructions (each a sequence of words in a pseudolanguage) and their corresponding output sequences (each a sequence of colored circles). After participants mastered the study instructions, they were provided with 10 novel (query) instructions and asked to generate the corresponding output sequences (4 queries are shown in Fig. \ref{fig_miniscan}B; the full set is shown in Extended Data Fig. \ref{fig-sysgen-methods}). Four words mapped to isolated output symbols (e.g., ``dax'' to RED and ``zup'' to YELLOW), and three words are functions that take one or two arguments (e.g., ``blicket'' is like ``surround'' so when you ``dax blicket zup'', the dax surrounds the zup, or RED YELLOW RED). Up to two function compositions were observed during training, and up to three function compositions were evaluated at test. Participants ($N=23$) were capable compositional learners such that on average 80.7\% of responses matched the rule-based symbolic system used to design the task, but nearly 20\% of responses followed other patterns. 

We examined how BBT can be used to better understand this complex, quasi-compositional mix of human behavior. We can specify a Bayesian model over sets of rules in the spirit of the other case studies. Following \cite{LakeMLC}, the prior $P(h)$ is defined through probabilistic context-free grammar, sampling hypotheses $h \in H$ such that each is a set of rewrite rules. For the likelihood $P(y_i | h; x_i)$, the output $y_i$ is the result of recursively applying the rules in $h$ to input $x_i$ with probability $1-\epsilon$, otherwise with probability $\epsilon$, the output $y_i$ is produced by uniformly sampling output symbols in sequence until the end-of-sequence symbol is drawn ($\epsilon=0.01$). After distillation on a set of 100,000 episodes, BBT was fine-tuned on examples of human behavior. For this purpose, we collected a new dataset of 166 unique participants distributed across eight different rule systems, with two shown in Extended Data Fig. \ref{fig-sysgen-methods}. Section \ref{methods_miniscan_finetuning} of the methods provides additional information about the Bayesian model and the human data for fine-tuning.

As in the previous case studies, we compared the neural networks trained with distillation only, fine-tuning only, and the full BBT pipeline. We did not implement a special-purpose approximate inference algorithm for the symbolic Bayesian model, instead relying on the approximation provided by model distillation. Each network was trained five times with different random seeds, using six rule systems for fine-tuning and two for validation. The best run was selected and, to make the most of the limited human data, fine-tuning from that run's distillation only network was conducted five more times with different seeds, for the same number of epochs as found during early stopping, although it was tuned on all eight rule systems instead of just six. Following \cite{LakeMLC}, we report predictions on the new task for the run that had the highest log-likelihood of the grammar-specified answers. 

The models' predictions can be compared through the log-likelihood of the human behavior on the instruction learning task (Table \ref{table_ll}). The predictions show that BBT's behavioral fine-tuning step provides a 122.8 gain in natural log points over the distillation only network. The gain over the fine-tuning only network is much larger. BBT also captures the kinds of mistakes that people make, including their characteristic patterns and heuristics. When they made errors, people tended to follow a one-to-one bias, mapping one input symbol to one output symbol (24.4\% of human errors; 44.4\% of BBT errors; labeled as 1-to-1 in Fig. \ref{fig_miniscan}). People also tended to follow an iconic concatenation bias, meaning they preferred to map first input argument to the first output position, rather than the reverse as function 3 requires (23.3\% of human errors and 7.0\% of BBT errors that involve function 3; labeled as IC in the Fig. \ref{fig_miniscan}). These patterns are not present in the original Bayesian model, or at least not to the same degree, as only 10.4\% of distillation only errors are involved the one-to-one bias and 0\% involved iconic concatenation. In fact, as shown in Fig. \ref{fig_miniscan}, the distillation only network makes far fewer errors in general, and when it does make errors, they are not as human-like as BBT.

Although these biases were observed in \cite{LakeMLC}, it is notable that with BBT, we did not need to hand-design the training procedure to teach them to the network. In fact, the BBT network shows slightly better performance than the model with hand-engineered biases, performing with a log-likelihood of $-343.6$ compared to $-349.2$ \cite{LakeMLC}.

\section*{Discussion}

We introduced Bayesian distillation with Behavioral Tuning (BBT), a hybrid approach to computational cognitive modeling that combines the strengths of Bayesian and connectionist traditions. To capture the strong inductive biases of Bayesian models while also retaining the flexibility of neural networks, BBT models follow a two-stage recipe: first, a neural network is trained on synthetic data sampled from a Bayesian model in order to mimic it, and second, the neural network is fine-tuned on human behavioral data (Fig. \ref{fig_bbt_method}). In practice, this recipe interpolates between the two traditions: the network is pre-trained to approximate a Bayesian model, and then it is tuned to the degree that best predicts the human behavior using an early-stopping criterion (Fig. \ref{fig_bbt_intro}).

% The case studies cover domains that are traditional challenges for neural networks. Now, not only can they keep up with Bayesian models, they can outperform them.
We applied BBT to four case studies spanning learning from very limited data (studies 1, 2, and 4), learning logical representations (studies 2 and 3), and learning compositional rules (study 4). Each of these domains has traditionally been characterized as challenging for neural networks \cite{Geman1992,Fodor1988,LakeBaroni2018, Lake2016} and as relative strengths for Bayesian models formulated over structured representations \cite{Tenenbaum2011,LakeScience2015,Piantadosi2016c}. These characterizations are beginning to change in light of the recent advances in neural networks \cite{WhitherSymbols}, including work on Bayesian distillation related to the current article \cite{Muller2022,LakeMLC, McCoy2023,jagadish2025meta,Marinescu2024,PFNNature}. However, it can be unclear what a distilled network offers beyond an efficient approximation of the Bayesian model it was distilled from. Our results show that neural networks can do more than mimic models; with fine-tuning, they can predict human behavior more accurately, even in domains thought to favor Bayesian models.

% BBT shows how to unlock training nets with millions of params from scratch on limited human data
It is notable that the BBT recipe can successfully train powerful, 30-million-parameter transformers for modeling a particular task or family of tasks. In case study 4, for example, the model was fine-tuned on data from just 166 human participants, each answering eight queries, in order to predict the behavior of novel participants on a novel task. Traditionally, training a model of that size on such limited human data would have been a non-starter, given the data-hungry reputation of neural networks \cite{Geman1992}.  However, we find that BBT can unlock new capabilities in neural networks by distilling before fine-tuning, echoing an approach from statistics where fitting and sampling from simpler parametric models bootstraps the fitting of more complex ones \cite{HuangCatalyticPriors}. Without this first stage and its synthetic data, the more powerful and flexible models can fit poorly, as we find with the ``fine-tuning only'' networks in our experiments (Table \ref{table_ll}).

% The need for distillation changes with more data
The impact of distillation will depend on the setting. If the behavioral dataset for a task family is very large, training can proceed directly without distillation \cite{Peterson2021,Agrawal2020,EcksteinNHB2026}. For example, Peterson et al. \cite{Peterson2021} trained a network with a few thousand parameters to fit human judgments on thousands of choice problems. The network showed superior performance and also guided the construction of more interpretable models. Additionally, a subset of the same authors proposed training general-purpose neural networks as upper bounds on the predictive performance of other models \cite{Agrawal2020}. We see BBT as aligned with this perspective. Distinctively, however, BBT allows for training larger models on more limited data and interpreting how the human behavior deviates from simpler accounts.

% Another trend has been using LLMs, including fine-tuning LLMs, to predict human behavior at a scale much larger than the one considered here.
If training larger models was the only goal, large language models (LLMs) dwarf the BBT models considered here, although with substantial tradeoffs. Off-the-shelf LLMs have been used to model human behavior by prompting them directly \cite{WebbAnalogy2023,BinzGPT3} or as a means of sampling Bayesian hypotheses \cite{Ellis2023a}, despite the uncontrolled and often unknown nature of their training data. LLMs have also been fine-tuned on human data to adapt them for cognitive modeling, including the prominent Centaur model \cite{CentaurNature} fine-tuned on data from many studies and an LLM fine-tuned on data from a single study, specifically from Piantadosi et al. \cite{Piantadosi2016c} as used in case study 2 \cite{Loo2026}. The Centaur approach offers some advantages over BBT: a single LLM can handle many tasks, LLMs have substantial background knowledge that shapes new learning (like people \cite{Murphy2002}), and LLMs do not require distillation from a Bayesian model. In contrast, Centaur's pre-training dataset is uncontrolled while BBT's training is fully controlled, making Centaur's successes harder to interpret. Others have critiqued Centaur for learning shortcuts in human response sequences, provided as in-context examples, rather than genuinely engaging with the tasks \cite{CentaurShortcuts}. BBT avoids these challenges by training the network to do the task during distillation and otherwise providing only the previous answers as in-context examples (and not the previous human responses).\footnote{Note, in case study 2, the model's auto-regressive means of responding to a set of query objects allows it to see (between 0 and 4) behavioral responses to the previous objects while responding to the next. However, no behavioral responses are shown for any of the previous sets / trials.} Furthermore, the controlled nature of the training pipeline allows the modeler to examine why the BBT model performs as it does and how it deviates from ideal Bayesian representations and behavior.

% BBT models can be interpreted
Compared to LLMs as base models for fine-tuning, the resulting BBT models need not be opaque. By comparing behavior and internal representations from before and after fine-tuning, we identified how BBT models depart from their initial Bayesian models. In case study 1, we found a stronger influence of mathematically simpler rules (evens, odds, ending in n) and a weaker influence of more complex rules (powers of n and multiples of n) when compared to the classic Bayesian model of this task \cite{Tenenbaum1999}. In case study 2, we found a similar shift from complex rules (with quantifiers) to simpler rules (without quantifiers, and even to trivial rules like constants). Likewise, in case study 3, we found an increase in proportions of simpler rules (Type-I hypothesis). Finally, in case study 4, we recover a set of characteristic biases that guide the human generalization in compositional instruction learning, consistent with past work that manually added these biases to the model training \cite{LakeMLC}. Together, these findings show how BBT offers not only improved predictions but also psychologically meaningful analyses of how human behavior strays from existing models.

% Limitations
% - modeler must decide how to visualize
% - stronger interpretability is possible
% to see if BBT can help adjuidcate between different hypothsized inducite biases
BBT also has important limitations. First, interpreting a fine-tuned network requires choices about which  episodes, layers, and variables to visualize, e.g., Fig. \ref{fig_ngame}. Second, some of our analyses, particularly the sparse approximation of model behavior (Section \ref{methods_sparse} of the Methods), are restricted to the hypotheses from the original Bayesian model and therefore characterize changes in the influence of those hypotheses. They may miss new hypotheses or changes to the logic of Bayesian inference that would better characterize human behavior, although auto-interpretation methods \cite{transcoder, rmus2026generating,jagadish2026closing} or  training the networks to show their reasoning could help address this limitation. Third, BBT introduces additional modeling choices concerning the Bayesian model used for distillation. In informal experiments, we found that broader priors and more deterministic likelihoods tended to lead to better BBT performance, perhaps because they encourage the network to learn the logical structure of the domain during training while leaving the noise and exceptions to be learned during fine-tuning. For example, although the best Bayesian model from Piantadosi et al. had likelihood parameters that modeled about half of the object labels as random noise \cite{Piantadosi2016c}, we trained BBT with no noise and instead relied on fine-tuning to adapt. Although the benefits of these modeling choices remain speculative and should be confirmed in more systematic experiments, we are hopeful that BBT can help to uncover the mental processes hiding behind these kinds of noise parameters.

% Future directions 
%  we did not try distilling different Bayesian modeling for the same task,
% random effects modeling. modeling individuals rather than aggregate.
% Is this approach relevant for non-Bayesian models, e.g., RL? parametric discriminative models?
Several extensions could broaden the scope and interpretability of BBT modeling. One direction is to distill competing Bayesian models and then compare their performance after identical fine-tuning. Such comparisons would help to broaden the search for the inductive biases that best explain human behavior. 
A second direction is to model individual differences rather than population behavior. In case study 3, we supplied an embedding indicating whether a participant was young or old, analogous to a fixed effect in a generalized linear model. Similarly, we could supply an embedding indicating which participant is responding, analogous to a random effect of participant. With a different learned embedding for each participant, the geometry of these embeddings could reveal continuous variation and/or cluster structure in their strategies. 
A third direction is to apply the general two-stage recipe to enrich cognitive models outside the Bayesian tradition, including reinforcement learning models (see related work on hybrid reinforcement learning models by \cite{EcksteinNHB2026} and \cite{LiuNeurIPS2025}), production systems \cite{HPS1972}, and other approaches constrained by their strong parametric or representational assumptions. 
A fourth direction is closing the loop between BBT and automated methods for interpreting its learned representation \cite{transcoder, choi2024automatic, paulo2024automatically}, potentially allowing the system to discover new symbolic hypotheses and then feed them back into the hypothesis space of the Bayesian model.
% This provides a data-driven way to discover new symbolic models of human behavior, unlike previous approaches that either rely on human hand-crafting \cite{ Agrawal2020} or LLM priors \cite{rmus2026generating, binz2025automated, Wong2025} .

% Conclusion -- how these biases develop
Finally, BBT raises questions about where the inductive biases represented in its models come from. The two stages of BBT training (Fig. \ref{fig_bbt_intro}) should not be interpreted as a developmental claim that children's minds start near Bayesian solutions and then adapt. A more developmentally oriented approach would need to rethink this recipe, starting with innate ingredients and inductive biases and then learning from experience. Specifically, a model could begin by distilling plausible innate structure into a neural network and then training on realistic egocentric input, such as headcam video from children \cite{SAYCam,Long2024,Vong2024Science}, rather than tuning on adult behavior. The present work has a more limited objective: to identify models that best predict how people learn new concepts. Across the four domains examined here, combining Bayesian modeling and connectionist fine-tuning produced more accurate behavioral predictions than either component would have alone and revealed psychologically meaningful departures from the original Bayesian accounts.

\section*{Methods}
\renewcommand{\thesubsection}{M.\arabic{subsection}}
\setcounter{subsection}{0}

\subsection{Implementation of BBT across the case studies} \label{methods_arch}
The starting point for the architecture, optimizer, and hyperparameters is based on  \cite{LakeMLC}, although a number of adjustments were made for the increase in scale from that work. The model was trained to minimize the cross-entropy loss (averaged over tokens) between model predictions and the target sequences. The two-stage training process is illustrated in Fig. \ref{fig_bbt_method}. In the distillation stage, the target sequences are synthetic data from the Bayesian model. In the fine-tuning stage, the target sequences are examples of human behavior. The architecture and the implementation of both stages are detailed below.

\textbf{Architecture.} The model $f_\theta(\yprime ; \xprime, X_m,Y_m)$ is a sequence-to-sequence transformer \cite{Vaswani2017} with about 30 million trainable parameters. This is scaled up from the architecture with 1.4 million parameters in \cite{LakeMLC}, which is also capable of learning the tasks in our case studies, although pilot simulations found that the larger model performs better. The model consists of two neural networks working together: an encoder that processes the input (concatenated $\xprime, X_m,Y_m$), and a decoder that predicts the output $\yprime$ given the encoded input. The encoder and decoder each have 4 layers, 8 attention heads, 512-dimensional embeddings, and a 2048-dimensional MLP hidden layer with GELU activation functions \cite{Hendrycks2020}. Dropout with probability 0.1 is applied to the input embeddings (after absolute sinusoidal position encoding) and to the transformer layers. Note that an encoder-decoder transformer was chosen to support bidirectional attention over the concatenated input variables; however, this is an implementation detail and a decoder-only architecture should work as well.

\textbf{Distillation.} The model was trained for 50 epochs using the AdamW optimizer \cite{Loshchilov2019} with a weight decay of 0.01, $\beta_1 =0.9$, and $\beta_2 = 0.95$. In contrast to the linear scheduler in \cite{LakeMLC}, the base learning rate was 0.0001, with linear warm-up during the first epoch followed by a reduce-on-plateau scheduler (by a factor of 10), which monitors the validation loss for stagnation with a patience of 4 epochs. (Note that a patience of 5 was used for the logical concepts.) Gradients were clipped to have a maximum norm of 1.0. The scheduler was allowed to decrease the learning rate 3 times; upon the 4th attempt, training was halted. The distillation stage for most models was done on a single NVIDIA L40 GPU, although a H200 GPU was used for the logical concepts and shepard category learning.

The datasets for distillation required sampling episodes from the Bayesian models. This was done as follows: for the number concepts, there were 100,000 episodes for training and 200 for validation; for logical concepts, 200,000 for training and 400 for validation; for Shepard concepts, 1,000,000 for training and 2,000 for validation; for compositional instruction learning, 100,000 for training and 200 for validation, using the same episodes as used for training the MLC (algebraic only) model in \cite{LakeMLC}. The validation episodes were only needed for the learning rate scheduler, and thus the episodes were not necessarily novel compared to training; for instruction learning, the validation episodes came from different grammars \cite{LakeMLC}.

The batch size was determined by the number of episodes fit on the GPU. A single batch is an aggregation of $k$ different training episodes. Each of these episodes consists of a study set of $m$ input-output pairs ($X_m$ and $Y_m$) and a query set with $n_q$ different queries. Thus, the effective number of query input-output pairs in a batch is $k*n_q$. The batching of episodes was done as follows: for the number concepts, $k=50$, $m \sim \text{Unif}(1,10)$, and $n_q=20$; for the logical concepts, $k=100$, $m \sim \text{Unif}(0,25)$, and $n_q=10$; for the Shepard concepts, $k=100$, $m \sim \text{Unif}(0,16)$, and $n_q=8$; for compositional instruction learning, $k=50$, $m \sim \text{Unif}(0,14)$, and $n_q=10$.

\textbf{Fine-tuning.} The fine-tuning stage mirrors the distillation stage except as noted below. After distillation, with weights saved from the final step, the network was fine-tuned on human behavioral data for 20 epochs (50 for Shepard category learning) using the same optimizer, scheduler, learning rate, gradient clipping, and other settings reinitialized as at the start of distillation, except that the scheduler patience was only 2 epochs. A variant of early stopping was implemented to select the stopping point adaptively and avoid overfitting, as illustrated in Fig. \ref{fig_bbt_intro}. Although fine-tuning was not actually stopped early, the validation loss was tracked across the 20 epochs at intervals of 100 steps, and the parameter values with the best validation loss across fine-tuning were saved. These saved values were used rather than the values at the final step.

The batching of episodes was done as follows: for the number concepts, $k=25$ and $1 \leq m \leq 4$; for the logical concepts, $k=10$ and $0 \leq m \leq 24$; for the Shepard concepts, $k=100$ and $1 \leq m \leq 15$; for compositional instruction learning, $k=10$ and $m=14$. Across the case studies, each episode consisted of a 
a single input query paired with $n_q$ responses from different human participants; thus, $n_q$ varied for each episode based on the number of available participants.

\subsection{Sparse approximation} \label{methods_sparse}
To identify the hypotheses consistent with a model’s predictions, we approximate the predictions for an episode with a sparse mixture of a few human-readable hypotheses drawn from the original Bayesian model. Our approximation is based on the posterior predictive distribution of a Bayesian model (Eq. \ref{eq_pp}). To see this, for simplicity,
consider hypotheses $h$ that assign a deterministic label $y \in \{0,1\}$ to a query object $x$
such that $y = h(x)$. Because $P(y = 1|h; x) = \mathbbm{1}[h(x) = 1]$ where $\mathbbm{1}$ is the indicator function, we can write the Bayesian model's posterior predictive (Eq. \ref{eq_pp}) as
\begin{equation}
P(\yprime = 1 | Y_m; \xprime,X_m) = \sum_h P(h|Y_m;X_m) \mathbbm{1}[h(\xprime) = 1],
\end{equation}
indicating that the probability of predicting $\yprime = 1$ is a sum of the weights (posterior probabilities) of all hypotheses that make this prediction.

Using the posterior predictive as an analogy, we approximate the output of the neural network with fitted weights $w_h$,
\begin{equation} \label{eq_sparse}
f_\theta(\yprime = 1; \xprime, X_m, Y_m) \approx (1-\gamma) (\sum_{h} w_h \mathbbm{1}[h(\xprime) = 1]) + \gamma 0.5,
\end{equation}
where the weights $w_h$ are positive, sum to 1, and are encouraged to be sparse for interpretability. The parameter $\gamma$ is a lapse rate that absorbs noise. For example, for the number game with study set \{8, 80, 48\}, BBT predictions can be approximated by weighting the hypotheses ``multiples of 8'' (weight 0.26), ``ending in 8'' (weight 0.24), ``evens'' (weight 0.17), and so on, as shown in Extended Data Fig. \ref{fig_ngame_8_80_48}.

Fitting the weights $w_h$ and $\gamma$ is formulated as the following optimization problem. 
Let $Q$ denote a varied set of queries $x_q$ (e.g., all integers from 1 through 100 for the number game). For shorthand, denote the left-hand side of Eq. \ref{eq_sparse} by $f_\text{net}(\xprime)$ and the right-hand side by $f_\text{sparse}(\xprime; w_h, \gamma)$.
The optimization problem is 
\begin{equation}
\underset{w_h,\gamma}{\operatorname{argmin}}\;
\frac{1}{|Q|}\sum_{x_q \in Q}
D_{\mathrm{KL}}\!\left[
f_{\text{net}}(x_q)
\,\|\, 
f_{\text{sparse}}(x_q;w_h,\gamma)
\right]
+ \eta H(w_h).
\end{equation}
with the constraints on $w_h$ enforced by re-parameterizing as unconstrained and applying a softmax. The Shannon entropy $H$ serves as a regularizer that encourages the solution to concentrate weight on as few hypotheses as possible. The tradeoff is controlled by $\eta$, which we found works well at 0.005. The overall objective is optimized with L-BFGS in PyTorch, alternating between fitting $w_h$ and $\gamma$, after discarding hypotheses that make identical predictions. We validated that the algorithm could effectively recover the ground truth hypotheses from a distilled neural network's behavior, as demonstrated in the next section.

\subsection{Visualizing and recovering hypotheses with sparse approximation} \label{methods_sparse_recovery}
\textbf{Number concept learning.} To generate Fig \ref{fig_ngame}D, the distillation only network and BBT were presented with 1600 synthetic episodes, half generated by mathematical hypotheses and half by interval hypotheses. These episodes contained between 2 and 5 study examples and were filtered so that the original Bayesian model could recover the right category (mathematical vs. interval) for interval hypotheses and the exact generating hypotheses for mathematical hypotheses, using maximum a posteriori (MAP) inference. (Because interval hypotheses are densely distributed, exact recovery is not possible.) To evaluate the sparse approximation algorithm, we tested whether it could recover the ground truth hypotheses when analyzing the distillation only model (Fig. \ref{fig_ngame}D left): the best-fitting hypotheses were recovered exactly for 97.1\% of the episodes produced by mathematical hypotheses, although as expected, only 1.9\% of the interval hypotheses were recovered exactly. A hypothesis was considered recovered if it received the largest fitted weight.

\textbf{Logical concept learning.} To generate Fig. \ref{fig_set_concepts}D, the distillation only network and BBT were presented with 792 synthetic episodes produced by relatively simple hypotheses spanning various classes. Each episode had 24 study examples and 20 query examples; in this case study, each example is a set of objects. To avoid degenerate cases, we filtered the episodes such that each had at least two study examples and two query examples had a positive label in their corresponding object set. We selected relatively simple instances of that class. For instance, hypotheses in the class ``and'' were just a single conjunction of two features, e.g., ``object x is red and square'' where x is the object under consideration. (The classes for ``or'', ``implies'', and ``iff'' were analogous.) Each hypothesis in the ``forall'' class was a single quantifier and relation, e.g., ``all objects are larger than object x'' or ``all objects have the same color as object x.''  Each hypothesis in the ``exists'' class was a single quantifier and relation, e.g., ``there exists another object with the same color as object x.'' (``Exists one or a fewer'' was analogous.) To evaluate the sparse approximation algorithm, we found that it could also recover the ground truth hypotheses with 99.2\% accuracy from the distillation only network, considering all other hypotheses in Fig. \ref{fig_set_concepts}D as distractors rather than the entire (unbounded) hypothesis space. 

\textbf{Shepard category learning.} To generate Fig.~\ref{fig_shepard}D, the
distillation only network and BBT were presented with 2560 synthetic episodes,
generated from all $2^8 = 256$ possible assignments of the eight objects to two
categories in the Shepard category learning task. For each category assignment,
10 episodes were sampled, randomizing the mapping of object features, category labels, and the order of the support examples. 
Each episode consisted of 16 support examples (each of the eight objects presented
twice) and 8 query examples (each object once). Each episode was colored by its
most likely hypothesis, inferred from the model's query predictions via sparse
approximation (Section~\ref{methods_sparse}): hypotheses matching Types~I--IV
were colored blue, orange, green, and red, respectively, and all other
hypotheses gray. BBT predictions were derived by conditioning on the young and
older age embeddings separately, in both cases conditioned on block~1; the
distillation only network has no block or age embeddings and was evaluated
unconditioned.

\subsection{Fitting the lapse rate} \label{methods_fitting}
If the length of the target output is known (e.g., the output is just one symbol that is either ``yes'' or ``no''), the probability of a participant producing each output symbol $s \in S$ is $P(s) = (1-\lambda)P_M(s) + \lambda \frac{1}{|S|}$ where $P_M$ is the model prediction before the lapse mechanism and $S$ is the set of possible outputs with cardinality $|S|$. The same lapse model can also be applied if the length of the output is not known, except in that case, the end-of-sequence token ($<$EOS$>$) is added to the set of possible tokens $S$. If the model has no prediction for a particular symbol (e.g., this symbol extends beyond the model's predicted output sequence), $P(s) = \frac{1}{|S|}$.

\subsection{Notation for number concept learning} \label{methods_bayes_num_game}
The Bayesian model of the number game differs from the other case studies in how the data are sampled. In the number game \cite{Tenenbaum1999}, an observed number $x_i$ is sampled from a given hypothesis, leading to a likelihood of the form  $P(x_i|h)$. (We omit labels $y_i$ since all observed numbers have label 1.) In the other case studies, the object/observation $x_i$ is assumed to come from a different process that is not modeled. Conditional on $x_i$, only its label $y_i$ is sampled from a hypothesis, leading to a likelihood of the form $P(y_i|h; x_i)$. The semicolon notation indicates that the model conditions on $x_i$ rather than modeling its distribution. The posterior predictive distribution and its neural network approximation $f_\theta(\cdot)$ therefore differs from Eq. \ref{eq_pp}:
\begin{equation} \label{eq_pp_ngame}
P(\yprime = 1 | X_m; \xprime)= \sum_h \mathbbm{1}[\xprime \in h] P(h|X_m) \approx f_\theta(\yprime = 1; X_m, \xprime).
\end{equation}
We do not need to adapt the neural network to accommodate these different sampling assumptions. The network instead learns the appropriate assumption through distillation: if the observations arise through an informative sampling process, the network will be incentivized to model that process to make more accurate predictions.

\subsection{Rational rules model} \label{methods_bayes_shepard}

% Data generation for distillation
% Each concept defines an episode over the eight objects formed by crossing three binary dimensions; the assignment of object features to language tokens (shape: triangle/rectangle, color: blue/yellow, size: size1/ size2) was randomized per episode. The support set contained each of the eight objects twice (16 items, shuffled), which was the length of a single block in Shepard concept learning tasks, and the query set contained all eight objects. Labels were produced by the concept's rule, with each label independently flipped to model outliers: a label is inconsistent with the rule with probability $\epsilon/2$, where $\epsilon = e^{-b}/(1 + e^{-b})$ and $b = 2$ (i.e., a flip probability of ${\approx}0.06$) or $b = 4.5$ (i.e., a flip probability of ${\approx}0.01$; see Extended Figures \ref{fig_shc_extended_2}). This outlier noise was applied to both support and query labels.

% \paragraph{Rational Rules model.} 
A total of $10^6$ hypotheses were drawn from rational rules prior over DNF grammar defined on three binary features, following \cite{Goodman2008a}, and evaluated each hypothesis on the eight stimuli. Production rules were kept fixed; Production probabilities were drawn independently for each hypothesis from a symmetric Dirichlet distribution with concentration parameter $\alpha = 1$, so that hypothesis generation is marginalized over production probabilities. The start rule requires at least one conjunction, excluding the trivial always-true concept.  We used Monte Carlo sampling to approximate the Rational Rules posterior. 
To make Bayesian inference tractable and fast, we collapsed hypotheses with identical extensions into a single hypothesis and the prior $P(h)$ of each hypothesis was set at its relative frequency among the $10^6$ samples; because shorter formulas are generated more often, this frequency-based prior implements simplicity bias. Given a support set, the likelihood followed the outlier model of \cite{Goodman2008a}: $P(y_i \mid h; x_i) = 1 - \epsilon/2$ if $h$ is consistent with the observed label and $\epsilon/2$ otherwise, with $\epsilon = e^{-b}/(1+e^{-b})$ and $b = 2~\text{or}~4.5$. The predictions for queries were computed from the posterior predictive dataset, smoothed by the same outlier probability, and combined with a lapse rate selected by grid search over $[0, 0.9]$ on the validation split of the respective behavioral dataset.

\subsection{Learning block and age embeddings for Shepard category learning} \label{methods_bayes_shepard_2}

% \paragraph{Learning block and age embeddings.} 
To capture how performance improves over experimental blocks and differs between age groups, we augmented the network $f_\theta(\cdot)$ with two learned memory embeddings during fine-tuning. 
We constructed fine-tuning episodes separately for each participant, category structure, block, and trial: the support set $(X_m, Y_m)$ contained the object stimuli and ground-truth (feedback) labels of the trials that the participant had already completed within the given block (i.e., between 1 and 15 trials), and the query, as before, was the participant's response to the subsequent trial.  

As the support set only carries trial-level information within a block, the neural network, by design, cannot infer which block the participant is completing or their age from the observed data. We therefore let the network learn memory embeddings related to these two features directly from the behavioral data: a block embedding with a 512-dimensional vector per experimental block ($b \in \{1,\dots,N\}$, where $N=6$ for the Badham et al. study and $N=12$ for the Lewandowsky et al. study), and an age embedding with one vector per age group (young or older). Given an episode's task block and the participants' age, the corresponding vectors are added (broadcast across positions) to the encoder output so that every encoded support representation carries the block and age signal before the decoder attends to it. We injected the embeddings after the encoding stage, rather than at its input, to prevent them from being attenuated by the encoder's layer normalization.

Both embeddings were initialized from $\mathcal{N}(0, 0.05^2)$; they were introduced and optimized jointly with the backbone during fine-tuning, using a separate parameter group with a learning rate of $10^{-3}$ so that freshly initialized embeddings could adapt faster than the pre-trained weights. The embedding table comprises eight 512-dimensional vectors (i.e., about $4,000$ parameters), and the architecture is otherwise unchanged. 
At inference time, the same fine-tuned network can be conditioned on any block--age combination.

% \paragraph{Lewandowksy dataset details.}  Lewandowsky \cite{Lewandowsky2011} conducted a study in which 113 adult participants performed all six category structures (Types~I--VI) over 12 blocks (12x16=192 trials)

\subsection{Compositional instruction learning} \label{methods_miniscan_finetuning}

\textbf{Interpretation grammars.}
In the Bayesian model of instruction learning \cite{LakeMLC}, a hypothesis to explain the input-output mappings is an ``interpretation grammar'', a set of rewrite rules for translating linguistic expressions into output sequences, inspired by formal semantics \cite{Geurts1999}. For example, the ground-truth hypotheses for the test episode and two fine-tuning episodes are shown in Extended Data Fig. \ref{fig-sysgen-methods} as panels A and B, respectively. Each episode has four ``primitive rules'' for translating an input symbol to an output symbol, e.g., in the test episode, \ll dax\rr $\rightarrow$ \red, \ll lug\rr $\rightarrow$ \blue, \ll wif\rr $\rightarrow$ \green, and \ll zup\rr $\rightarrow$ \yellow. The double brackets (\ll \rr) denote the interpretation function. Each episode also has three ``function rules'' that specify functions with variables (Extended Data Fig. \ref{fig-sysgen-methods}), with variables $u_i$ applying only to input primitives (e.g., ``dax'', ``lug'', ``wif'', and ``zup'' in the test episode) and variables $x_i$ applying to arbitrary non-empty strings. The interpretation process applies recursively, using the rewrite rules on intermediate expressions, until the expression is fully interpreted, i.e., has only output symbols and no double brackets. 

To see a worked example, let's interpret the linguistic expression ``dax fep'' according to the rules of the test episode by applying the interpretation function, \ll dax fep\rr. Of all the rules in the grammar, the rule \ll $u_1$ fep\rr $\rightarrow$ \ll $u_1$\rr \ll $u_1$\rr \ll $u_1$\rr\ applies because ``dax'' is a value that the variable $u_i$ can take. By applying this rule, the expression \ll dax fep\rr\ is rewritten as \ll dax\rr \ll dax\rr \ll dax\rr. The interpretation function is then applied recursively to the three intermediate expressions \ll dax\rr, which results in three applications of the rule \ll dax\rr $\rightarrow$ \red. The final output string is \red \red \red. An additional example is provided in Extended Data Fig. 3 of the original article \cite{LakeMLC}.

\textbf{Prior over interpretation grammars.}
Following \cite{LakeMLC}, the prior $P(h)$ is defined over interpretation grammars that have 4 primitive rules and 3 function rules, plus a concatenation rule. There are 8 possible input symbols and 6 possible output symbols. To sample a hypothesis $h$, first the primitive rules are sampled by choosing 4 input symbols uniformly at random and four 4 output symbols uniformly at random, both without replacement, and pairing them. Second, the function rules are sampled, with the left-hand side following either a one- or two-argument template, \ll $v_1$ input\_symbol\rr\ versus \ll $v_1$ input\_symbol $v_2$\rr, chosen by a flip of a fair coin. The placeholder ``input\_symbol'' is sampled uniformly from the remaining  options, after input symbols are chosen for the previous primitive and function rules. The template arguments $v_i$ are sampled uniformly without replacement from the set $\{u_1,u_2,x_1,x_2\}$. For each function rule, the right-hand side is a string of elements from the set \{\ll $u_1$\rr,\ll $u_2$\rr,\ll $x_1$\rr,\ll $x_2$\rr\}, although restricted to the same set of variables that appear on the left-hand side of that rule. The minimum right-hand side string length is 2 variables and the maximum is 8. For each variable slot after 2, there is a probability of 0.6 that the right-hand string terminates. Otherwise, the variable is sampled uniformly with replacement from the possible options. There is a final rule added that instantiates concatenation, \ll $u_1$ $x_1$\rr $\rightarrow$ \ll $u_1$\rr \ll $x_1$\rr, completing the sampling process for an interpretation grammar.

\textbf{Human data collection.}
Human participants in the United States ($N=226$) were recruited on Prolific for an experiment developed and run on the Smile platform (\url{https://smile.gureckislab.org/}). The study was approved by the Princeton IRB, protocol \#19605, and the participants were paid \$10 for their time. The participants were randomly assigned to one of eight episodes from different grammars, each with 14 study examples and 10 queries, excluding catch trials. These eight episodes/grammars were chosen because each of the distillation only networks (across all 5 random seeds) scored 100\% correct on the queries. Two example episodes are shown in Extended Data Fig. \ref{fig-sysgen-methods}B.

The experimental procedure from \cite{LakeMLC} was followed, with the full details available in the first Methods section of that article. In summary, the learning task was structured as a curriculum of 4 study phases, with the first 3 corresponding to different function rules. In each study phase, the study instructions included the 4 primitives and two examples of the relevant function rule. Participants were quizzed on their memory of the study instructions, repeating the quiz if they made a mistake. Each study phase was followed by a test phase, asking participants to generalize to novel queries by producing their outputs, as related to the relevant rule for that phase. Finally, there was a fourth study phase with all 14 study instructions shown together. The final test queries required engaging multiple function rules to answer correctly. 

We used the same procedure as \cite{LakeMLC} to ensure participants were engaged, including an instructions quiz, study phase quizzes, and catch trials where queries were the same as the study items. Participants ($N=60$) were excluded for the following reasons: 11 reported using external aids such as screenshots, pencil and paper, and/or AI assistants; 28 did not pass the study phases; 20 missed two or more catch trials; and 1 experienced technical issues. After exclusions, data from 166 participants were used to fine-tune the BBT model.

\subsubsection*{Acknowledgments}
We are grateful for comments on previous versions of this manuscript from Jonathan Cohen, Suyog Chandramouli, and Gaia Molinaro. We thank Changho Shin for suggesting the illustration in Figure 2.

\nolinenumbers
\baselineskip12pt
\bibliography{library_brenden_freeze,extra_refs}
\bibliographystyle{naturemag}

\clearpage

\renewcommand{\figurename}{Extended Data Figure}
\renewcommand{\tablename}{Extended Data Table}
\setcounter{figure}{0}
\setcounter{table}{0}

\begin{figure}[tb]
\centering
\includegraphics[width=.8\linewidth]{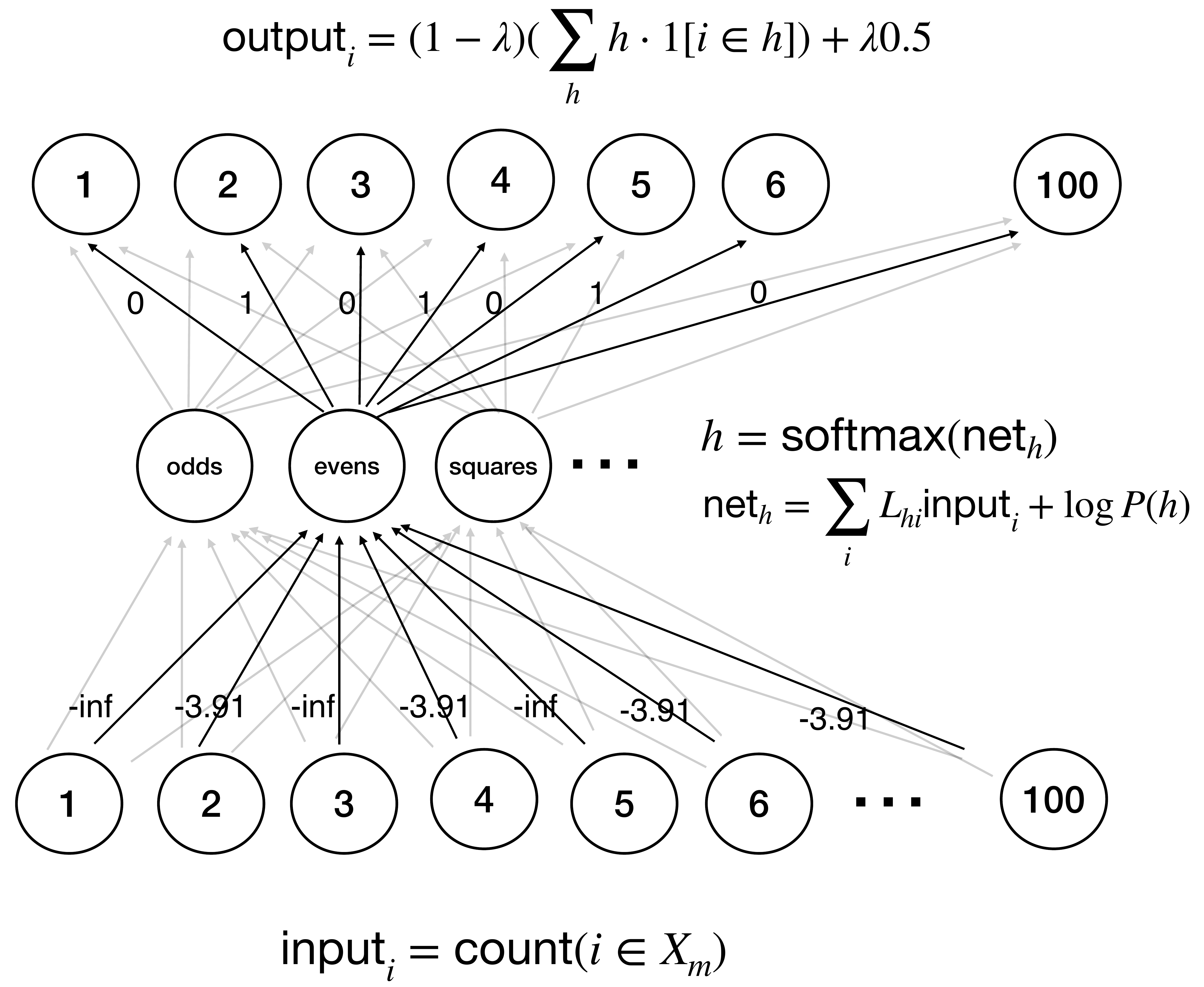}
\caption{Interpreting the Bayesian model of the number game as a multilayer perceptron. \textbf{Input layer}: The study set $X_m$ is represented as a count vector whose $i$th entry records the number of times integer $i$ appears in $X_m$, similar to a ``bag of words'' input vector. For instance, if ``4'' appears twice in $X_m$, $\text{input}_4 = 2$. \textbf{Hidden layer}: The activation of each hidden unit $h$ equals the posterior probability of the corresponding hypothesis, $P(h|X_m)$. The net input to the unit corresponding to each hypothesis ($\text{net}_h$) is the dot product between the input vector and weights $L_{h,:}$ that reflect the log-likelihood, $L_{h,i} = \log \frac{1}{|h|}$ if $i \in h$, otherwise $-\infty$. The weights $L_{h,:}$ for the hypothesis ``evens'' are shown on the arrows. The biases for the hidden units are the log prior, $\log P(h)$. The layer has a softmax activation function. \textbf{Output}: The network simultaneously outputs the posterior predictive probability for every possible query $i$, $\text{output}_i = (1-\lambda)P(i \in h | X_m) + \lambda0.5$ where the lapse rate $\lambda$ is optimized. \textbf{Prior-fit training}: For behavioral fine-tuning, the network is initialized at the original Bayesian model parameters. Only the 5084 prior logits corresponding to $P(h)$ and $\lambda$ are trained with L-BFGS, maximizing the log-likelihood of the human data with a regularizer (0.001) that encourages a higher-entropy prior. \textbf{Full-fit training}: This model extends the prior-fit model by including the log-likelihood matrix $L \in \mathbb{R}^{5084 \times 101}$ trainable. Each row is parameterized by unconstrained logits and normalized using log-softmax, using softened variants of their Bayesian model values for initialization. The indicator function in the output layer is relaxed to reflect that membership is now graded rather than binary, replacing $\mathbbm{1}[i \in h]$ with $e^{L_{h,i}} / (\max_j e^{L_{h,j}}).$ The code for training these models is released as well.}
\label{fig_ngame_alt}
\end{figure}

\begin{figure}[tb]
\centering
\includegraphics[width=\linewidth]{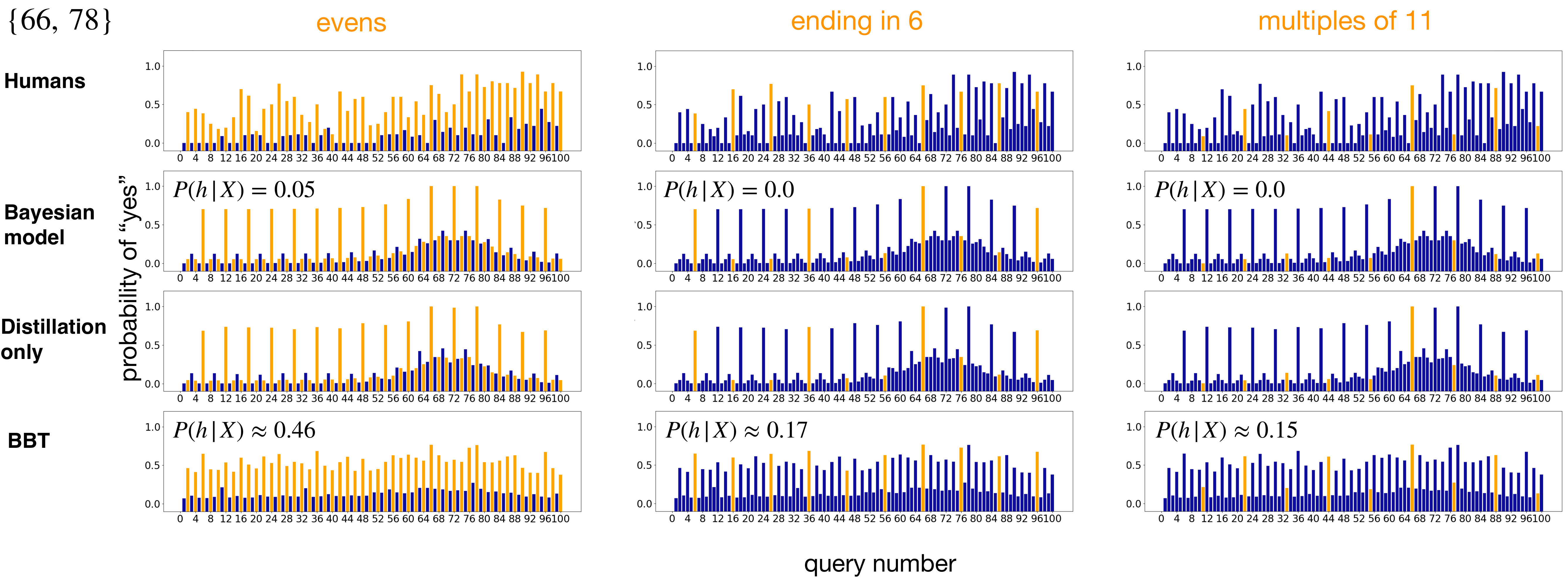}
\caption{Analysis of number game predictions given a program that accepts $\{66,78\}$. The proportion of participants to answer ``yes'' for each query 1 to 100 is shown, along with model predictions. Using sparse hypothesis decomposition for interpretation, the three hypotheses with the largest influence in BBT's predictions are shown, left to right across panels. The numbers included in each hypothesis are highlighted in orange.}
\label{fig_ngame_66_78}
\end{figure}

\begin{figure}[tb]
\centering
\includegraphics[width=\linewidth]{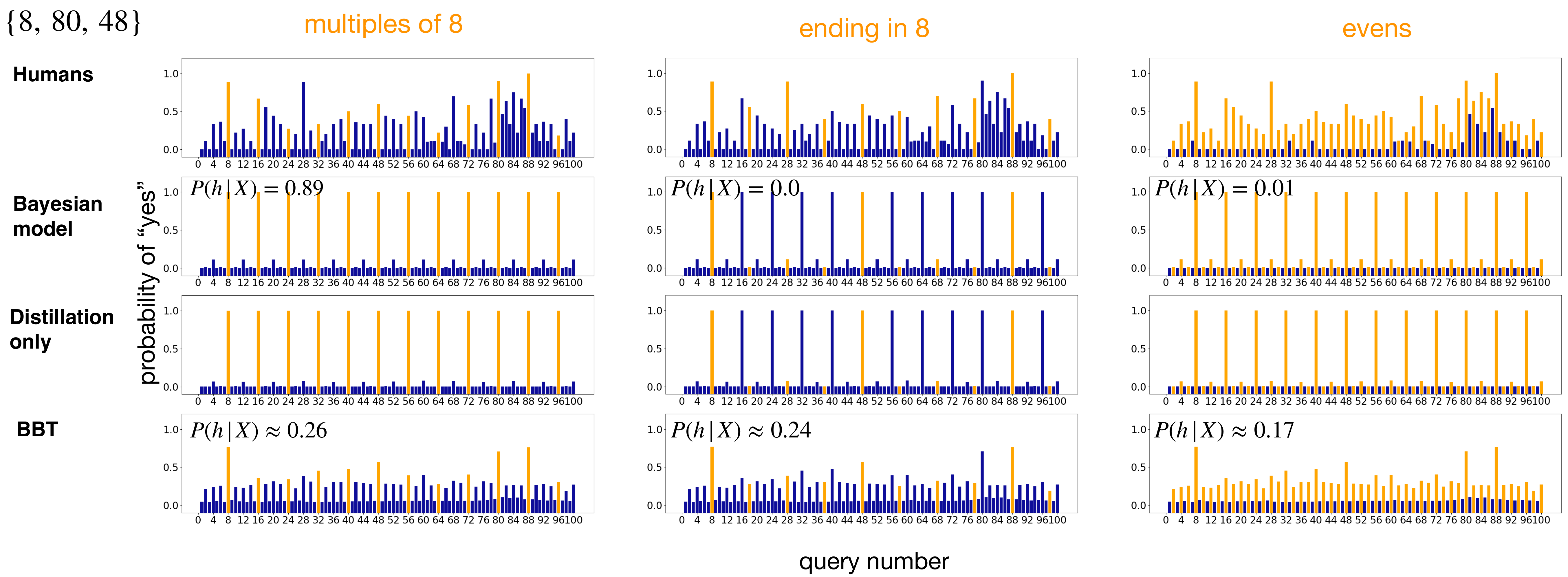}
\caption{Analysis of number game predictions given a program that accepts $\{8,80,48\}$. Otherwise refer to the caption in Extended Data Fig. \ref{fig_ngame_66_78}.}
\label{fig_ngame_8_80_48}
\end{figure}

\begin{figure}[tb]
\centering
\includegraphics[width=\linewidth,height=0.8\textheight,keepaspectratio]{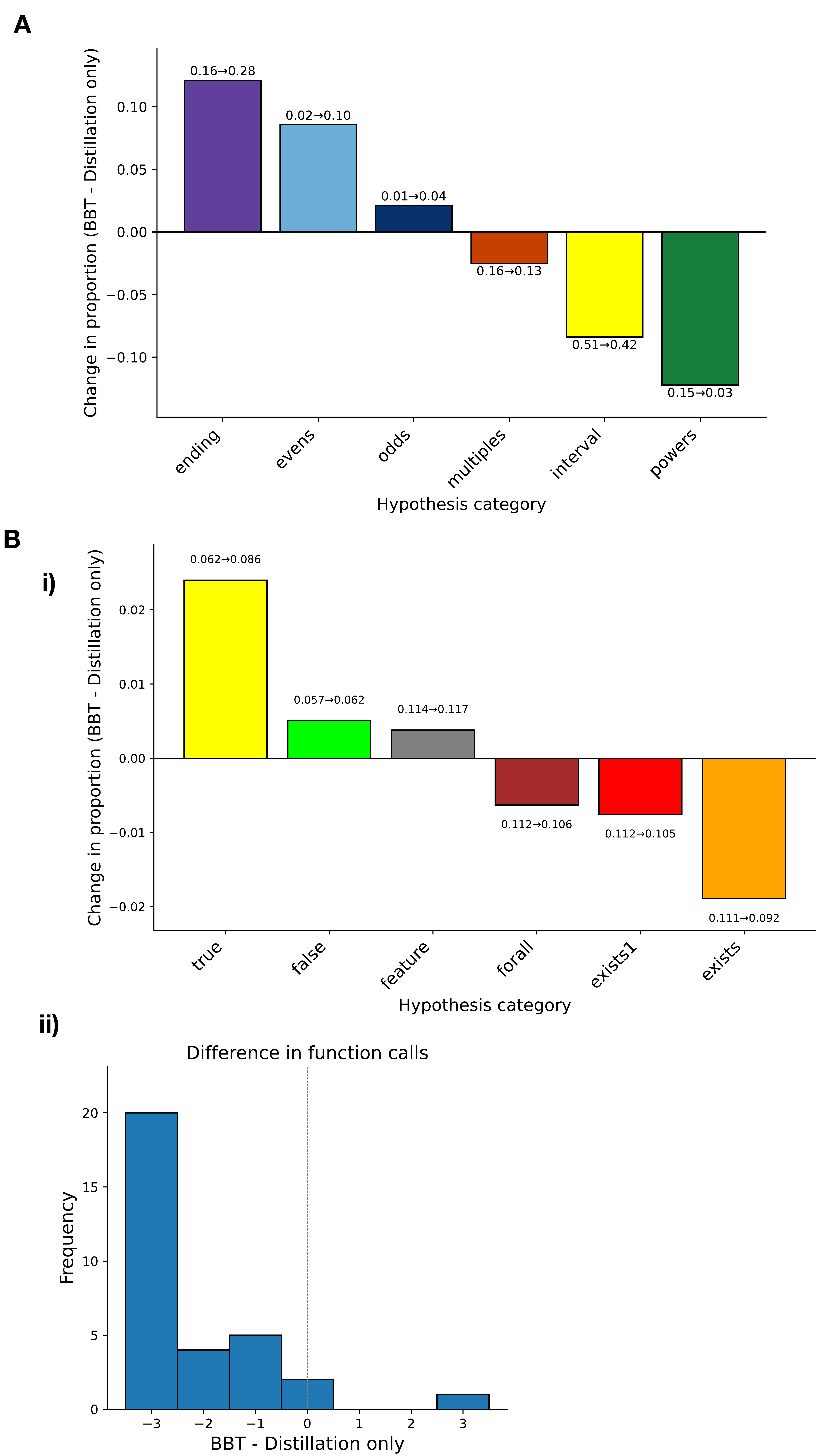}
\caption{Analysis of how hypotheses change between the distillation and fine-tuning stage. A) For the number game, the episodes in Fig. \ref{fig_ngame}D are categorized by their most likely hypothesis via sparse approximation (Section \ref{methods_sparse} of the Methods) for the two models, BBT and Distillation only. The difference in frequency (BBT - Distillation only) is shown on the y-axis. B-i) This is the same analysis for learning logical concepts based on the episodes in Fig. \ref{fig_set_concepts}D). Note that the following classes are not included for having an absolute difference less than 0.001: and, or, implies, and if and only if. B-ii) Histogram of the difference in function calls in the hypothesis expressions, showing that fine-tuning leads to fewer function calls, defined as the count of open parentheses.}
\label{fig_complexity_diff}
\end{figure}

\begin{table}[tb]
\centering
\begin{tabular}{l||lll}
Run & BBT & Distillation only & Difference \\
\hline
1 & -7,929.6 & -8,411.3 & 481.7 \\
2 & -7,616.9 & -8,235.0 & 618.1 \\
3 & -7,759.5 & -8,292.6 & 533.1 \\
4 & -7,769.2 & -8,376.9 & 607.7 \\
5 & -7,522.5 & -8,092.1 & 569.6
\end{tabular}
\caption{Predicting human behavior for novel logical concepts. Bayesian distillation was run five times, and each of these network runs was subsequently fine-tuned with BBT. Performance is reported as the overall log-likelihood as well as the difference in log-likelihood between the models (BBT - Bayesian distillation). All models have fit lapse rates (Section \ref{methods_fitting} of the Methods). }
\label{table_ll_logical_novel}
\end{table}

\begin{figure}[tb]
\centering
\includegraphics[width=\linewidth]{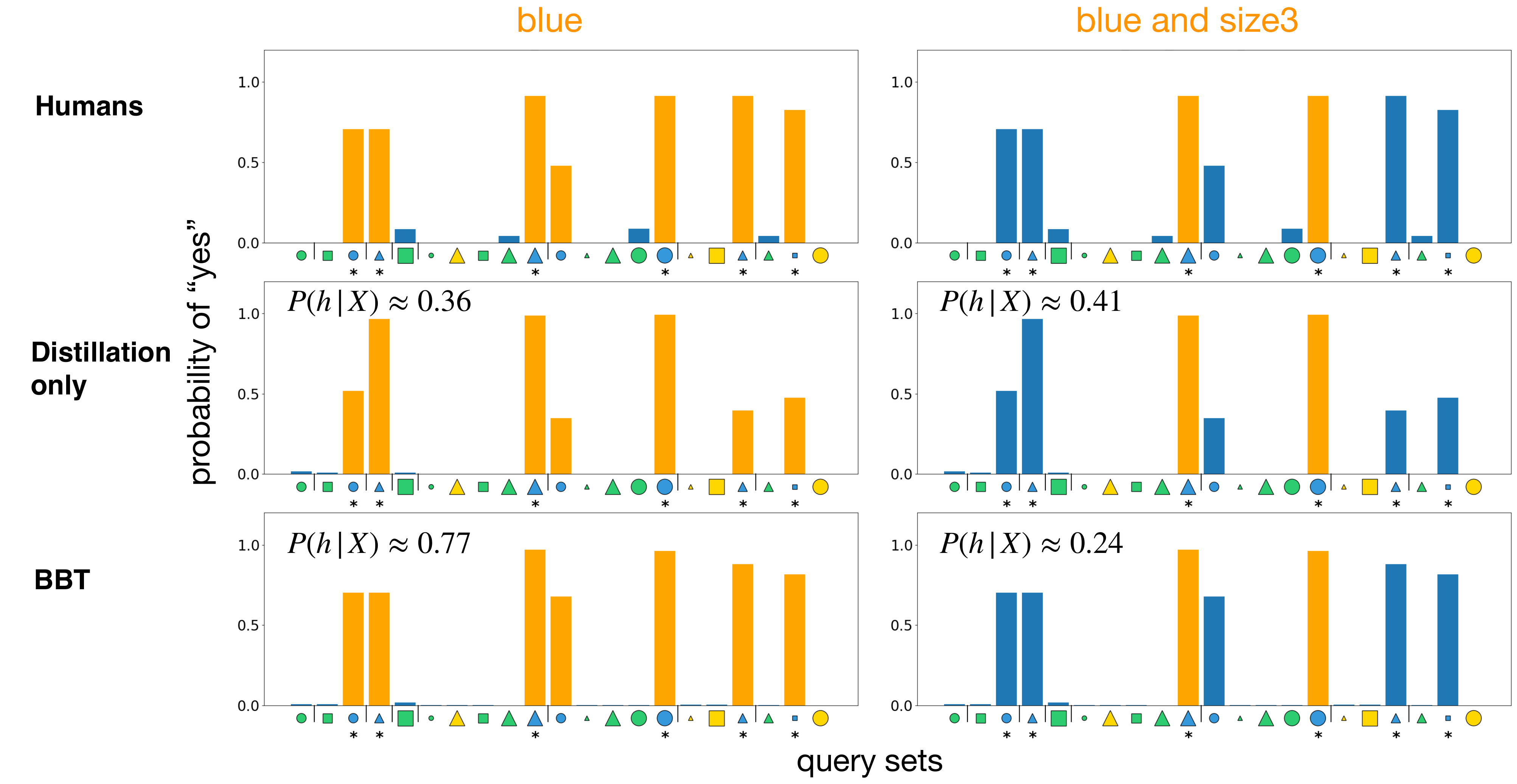}
\caption{Analysis of logical concept learning for the concept ``largest blue object in the set.'' The proportion of participants to answer ``yes'' for each query object is shown, along with model predictions. Using sparse hypothesis decomposition, the two hypotheses with the largest influence in BBT's predictions are shown, left to right across panels. The objects included in each hypothesis are highlighted in orange.}
\label{fig_set_is_largest_blue}
\end{figure}

\begin{figure}[tb]
\centering
\includegraphics[width=\linewidth]{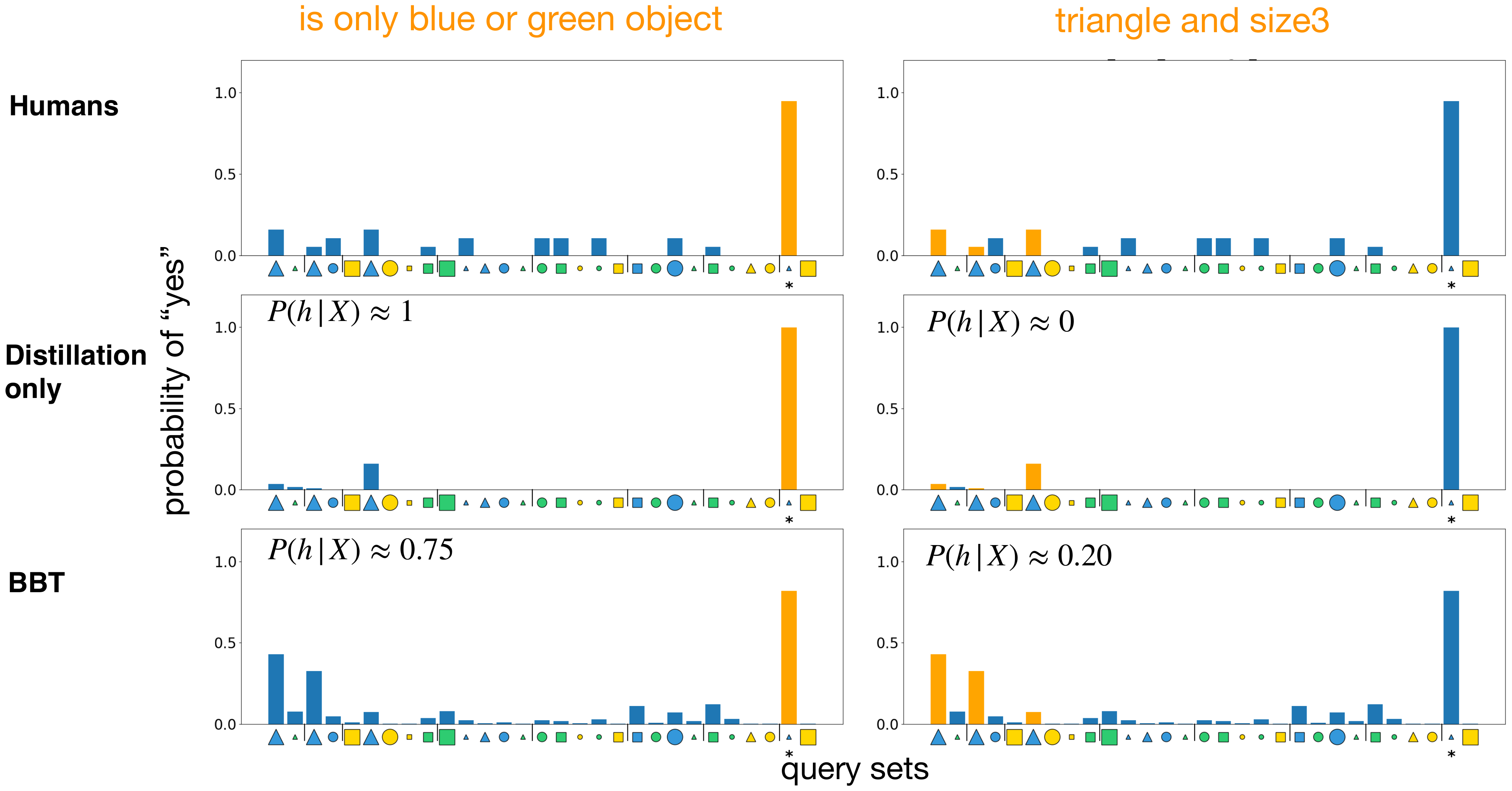}
\caption{Analysis of logical concept learning for the concept ``is the only blue or green object.'' Otherwise refer to the caption in Extended Data Fig. \ref{fig_set_is_largest_blue}.}
\label{fig_set_is_only_blue_green}
\end{figure}

\begin{figure}[tb]
\centering
\includegraphics[width=\linewidth,height=0.78\textheight,keepaspectratio]{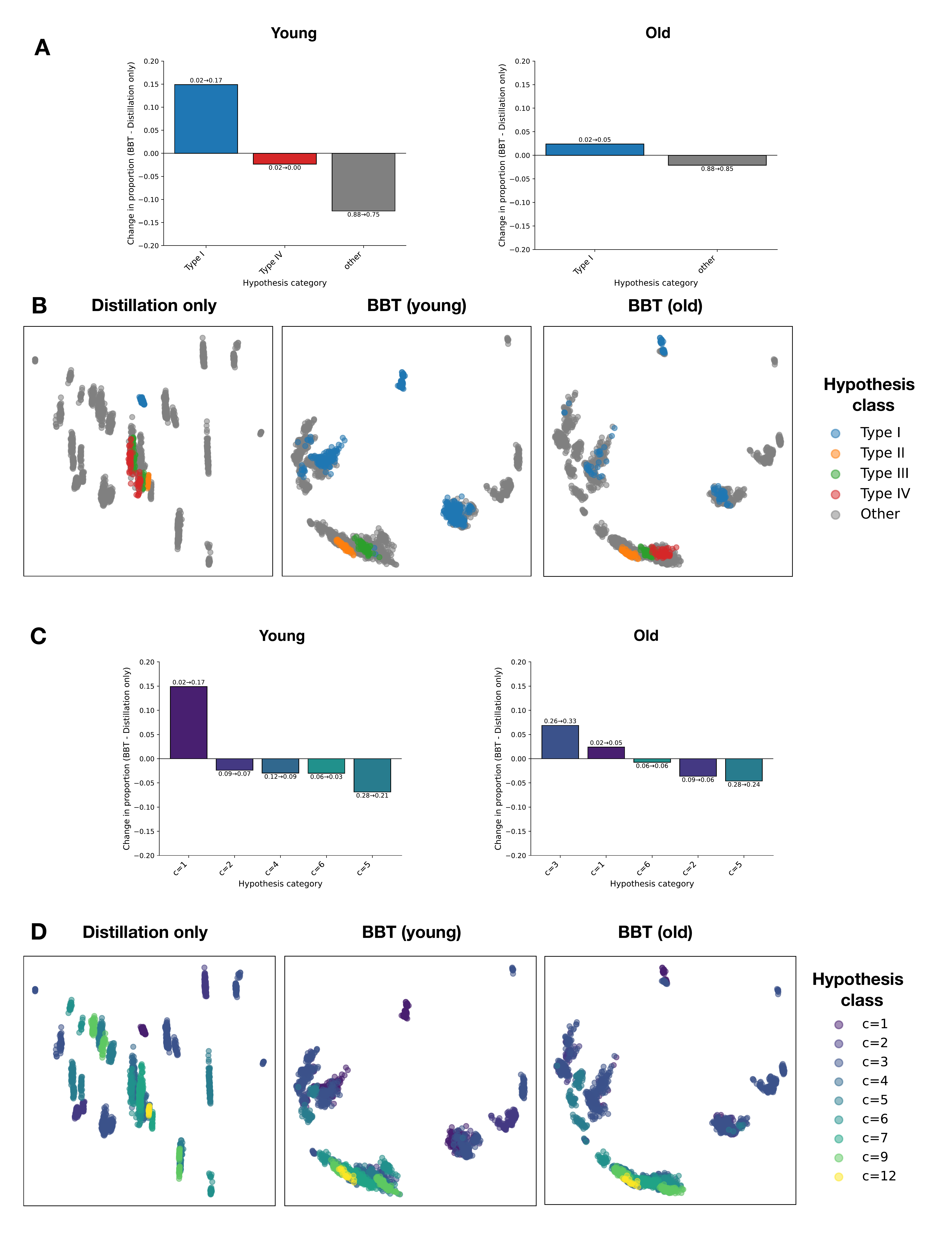}
\caption{ Analysis of how hypotheses change between the distillation and fine-tuning stage. For Shepard category learning, the episodes in (B and D) are categorized by their most likely hypothesis via sparse approximation for the two models (Section \ref{methods_sparse} of the Methods), BBT (young on the left and old on the right) and Distillation only. The results for BBT are shown separately into young and old participants by conditioning on age-specific embedding. PCA of model embeddings for distillation only (left) and human BBT (right) extracted from decoder layer 3. Each point is a different learning episode containing 16 objects in support and all unique objects as query, marked with the color of its (B) most likely hypothesis or (D) number of literals in the rule sampled from the probabilistic context-free grammar as fit to the model judgments via sparse approximation. The difference in frequency (BBT - Distillation only) of different hypothesis (A) and number of literals in a given hypothesis (C), whenever there was a difference observed in the sparse approximation analysis.}
\label{fig_shc_extended_1}
\end{figure}

\begin{figure}[tb]
\centering
\includegraphics[width=\linewidth,height=0.8\textheight,keepaspectratio]{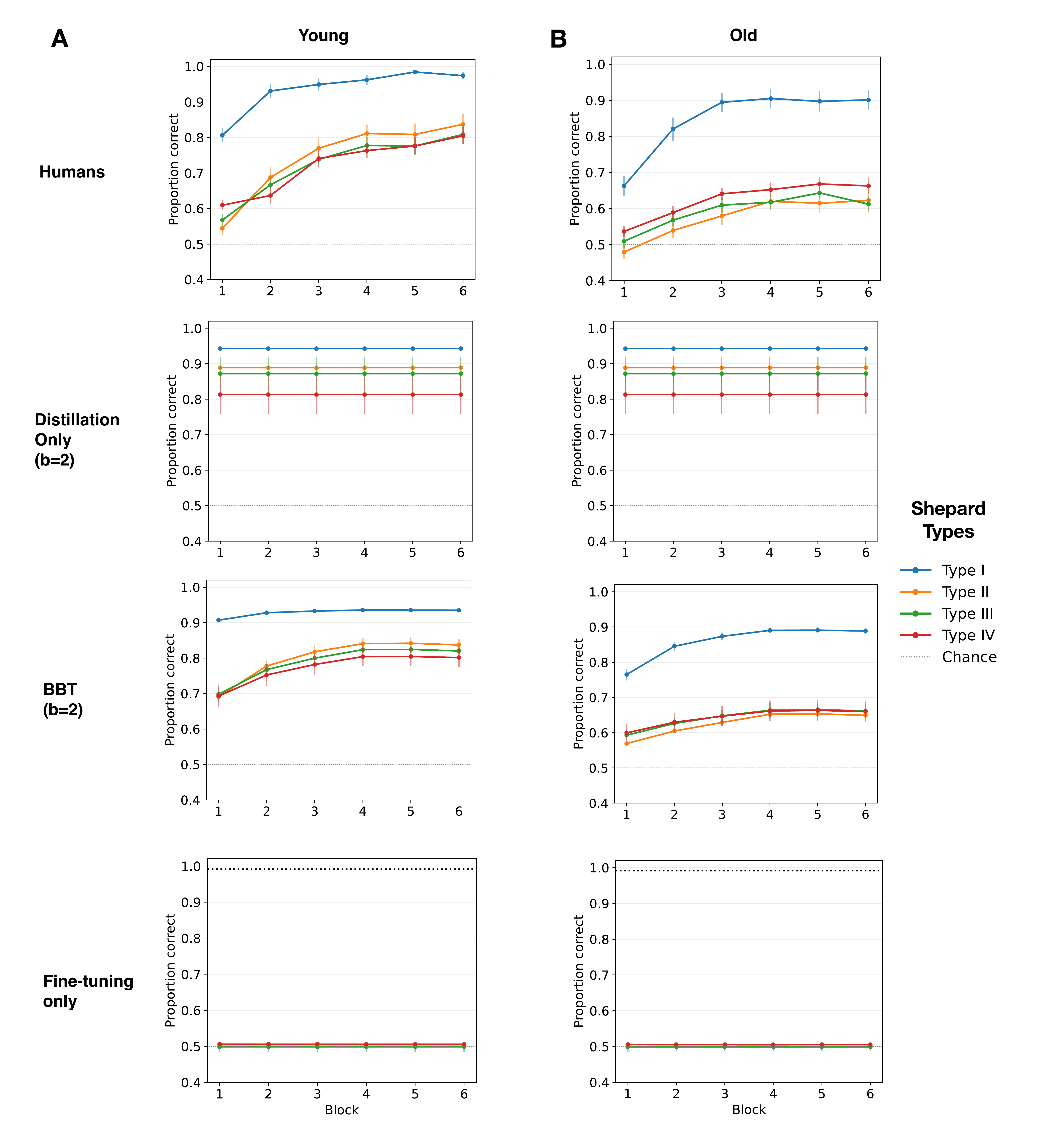}
\caption{A \& B) Learning curve over blocks for humans, distillation-only, and BBT for young participants (A) and older participants (B), data taken from the Badham et al. \cite{Badham2017} study. The y-axis shows the proportion of correct choices and x-axis is tasks blocks, with the four Shepard category structures (Type-I to Type-IV) shown as distinct lines. The distillation was done from a rational rules model with outlier probability set to 0.12 (i.e. $b=2$). As a result, the distillation only model, unlike the one used in main figure which was rational rules model with outlier probability set to 0.01, displays a clear ordering in difficulty (Type~I<Type~II<Type~III<Type~IV) when simulated on synthetically generated Shepard tasks. In addition, BBT ($b=2$) can reproduce the exact ordering of learning difficulties observed in humans, including the swap between type IV and type II observed in older participants, unlike distillation-only model and fine-tuning only model. }
\label{fig_shc_extended_2}
\end{figure}

\begin{figure}
  \centering
   \includegraphics[width=\linewidth,height=0.88\textheight,keepaspectratio]{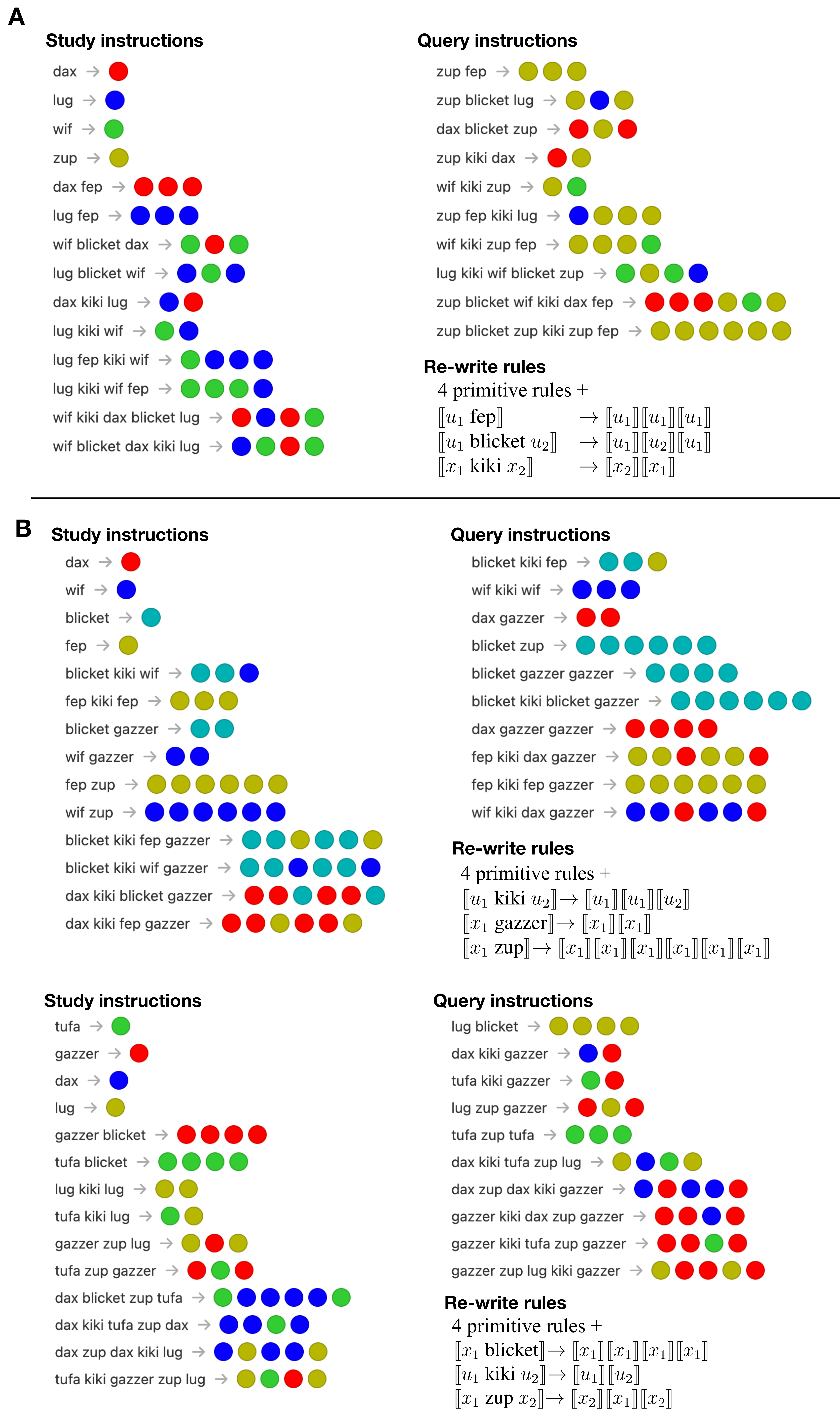}
  \caption{Three episodes for the compositional instruction learning task, including the test episode (A) and two fine-tuning episodes (B). Human participants and models are provided with the study instructions, which map linguistic expressions (pseudowords) to output sequences (color circles). The task is to produce the output sequences for the query instructions (the outputs are shown here, but not to participants). Each episode is generated by a grammar of rewrite rules, as explained in Section \ref{methods_miniscan_finetuning}.}
    \label{fig-sysgen-methods}
\end{figure}

\end{document}